\documentclass{article}
\usepackage{microtype}
\usepackage{graphicx}
\usepackage{subfigure}
\usepackage{booktabs}
\usepackage{algorithm}
\usepackage{algpseudocode}
\usepackage{enumitem}

\usepackage{tikz}       % Must be loaded in the main file
\usepackage{standalone} % Handles the external document formatting
\usepackage{tcolorbox} % For colored boxes
\usepackage[T1]{fontenc}
\usepackage{lmodern}
\usepackage{amsmath,amssymb}
\usepackage{xcolor}
\usetikzlibrary{arrows.meta}
\usepackage{amsthm}

\usepackage{amsthm}
\usepackage{multirow}
\usepackage{wrapfig}
\newtheorem{lemma}{Lemma}
\usepackage{xurl}
\usepackage{hyperref}

\usepackage[arxiv]{mlsys2025}
\usepackage{stfloats}

\mlsystitlerunning{Stealing Exact Training Data from Other Clients during Federated Learning}

\definecolor{ForestGreen}{cmyk}{0.864, 0.0, 0.429, 0.396}
\newif\ifcomments
\commentstrue

\begin{document}

\twocolumn[
\mlsystitle{Stealing Exact Training Data \\ from Other Clients during Federated Learning}

% It is OKAY to include author information, even for blind
% submissions: the style file will automatically remove it for you
% unless you've provided the [accepted] option to the mlsys2025
% package.

% List of affiliations: The first argument should be a (short)
% identifier you will use later to specify author affiliations
% Academic affiliations should list Department, University, City, Region, Country
% Industry affiliations should list Company, City, Region, Country

% You can specify symbols, otherwise they are numbered in order.
% Ideally, you should not use this facility. Affiliations will be numbered
% in order of appearance and this is the preferred way.
\mlsyssetsymbol{equal}{*}

\begin{mlsysauthorlist}
\mlsysauthor{Omkar Shrikanth}{inst1,equal}
\mlsysauthor{Soham Dambalkar}{inst1,equal}
\mlsysauthor{Parag Jyoti Pal}{inst1,equal}
\mlsysauthor{Advik Raj Basani}{inst1,inst2}
\mlsysauthor{Arnab K. Paul}{inst1}
\end{mlsysauthorlist}

\mlsysaffiliation{inst1}{Birla Institute of Technology and Science, Goa}
\mlsysaffiliation{inst2}{MATS Research}

\mlsyscorrespondingauthor{Omkar Shrikanth}{f20231120@goa.bits-pilani.ac.in}
% \mlsysaffiliation{anon}{Anonymous Institution}

% \mlsyscorrespondingauthor{Anonymous Author}{anonymous@example.com}

% You may provide any keywords that you
% find helpful for describing your paper; these are used to populate
% the "keywords" metadata in the PDF but will not be shown in the document
\mlsyskeywords{Federated Learning, Gradient Inversion, Language Models}

\vskip 0.3in

%\begin{abstract}
%In Federated Learning, Malicious entities can employ gradient inversion attacks to reverse engineer the private data using the updates of a federated learning round. While there have been developents in 
%A client-side attack model for exact text recovery in federated learning (FL) has so far been infeasible due to the high levels of noise introduced by the aggregation of gradients. We lead with our insight that recovery need not be done in a single shot and introduce a novel attack architecture designed as an iterative loop, in which the information from exactly recovered sentences of one round is used to denoise the aggregate gradient and enable further recovery in the next iteration. Posing as an honest-but-curious client in an FL setting, we demonstrate our attack pipeline on parallel-residual language models across three datasets. We show, for the first time, exact sentence recovery, with 100\% precision, and on aggregates containing up to $4\times$ more tokens than the model's hidden dimension, we recover up to 80.2\% of client sentences exactly, a regime where existing attacks recover none.
%\end{abstract}

\begin{abstract}
Malicious adversaries can employ gradient inversion attacks to reverse-engineer private data using the model updates of a round in Federated Learning. While there have been developments in gradient inversion for text, exact text recovery by a malicious client has so far remained infeasible, as \textsc{FedAvg} entangles the gradient contributions of many clients and obscures the signal of any individual sentence. We introduce a novel attack pipeline that unwinds this entanglement progressively through an iterative loop, in which the gradient contributions of sentences recovered exactly in one round are subtracted from the aggregate, exposing the remaining sentences for recovery in the next iteration. We show, for the first time, exact sentence recovery, with 100\% precision, and on aggregates containing up to $4\times$ more tokens than the model's hidden dimension, we recover up to 80.2\% of client sentences exactly, a regime where existing attacks recover none exactly.
\end{abstract}
]

% this must go after the closing bracket ] following \twocolumn[ ...

% This command actually creates the footnote in the first column
% listing the affiliations and the copyright notice.
% The command takes one argument, which is text to display at the start of the footnote.
% The \mlsysEqualContribution command is standard text for equal contribution.
% Remove it (just {}) if you do not need this facility.

\printAffiliationsAndNotice{\mlsysEqualContribution}  % leave blank if no need to mention equal contribution
%\printAffiliationsAndNotice{\mlsysEqualContribution} % otherwise use the standard text.

\section{Introduction}
\label{sec:intro}
Language models are increasingly finding applications in sensitive domains such as healthcare, finance, and law \cite{thirunavukarasu2023large, langinfinance, dehghani2025largelanguagemodelslegal}. Constructing such specialized models, however, typically requires fine-tuning, and in these settings, the training data often tend to be sensitive \cite{wu2023bloomberggptlargelanguagemodel, chen2024tmlr-survey}, comprising items such as medical records, financial reports, legal documents, and memos.
  
Federated learning (FL) \cite{mcmahan2017fedavg} is a natural fit for such scenarios \cite{yeopenfedllm}, enabling multiple parties holding domain-specific private data to jointly fine-tune a model without sharing their private corpora with one another or with a central third-party aggregator. FL works by having individual entities (clients) perform fine-tuning locally on their own data, transmitting only the resulting gradients to a designated aggregator that computes the model updates. The key insight is that, because training requires only the aggregation of gradients, individual entities can forgo sharing their private data altogether while still collaboratively fine-tuning a stronger model together.

\looseness = -1
However, the idea that the gradients offer no private information to a malicious adversary has generally been disproven, especially in the vision domain \cite{geiping2020inverting}.
% \cite{zhu2019dlg, zhao2020idlg} and language modeling \cite{tag, petrov2024dager,balunovic2022lamp} domains. While FL removes the need to share raw data, the gradients a client transmits to the aggregator have been shown to carry information about the underlying data. 
% A malicious entity can exploit this leakage to derive insights about or even reconstruct a client's original private training data. \anote{I'd suggest having some information about FedSGD / FedAvg here.}
The training and fine-tuning of visual classifiers has been shown to be invertible through relatively simple optimization-based methods. This is made possible largely by the continuous nature of the image data and the relatively small set of labels involved in such training scenarios \cite{zhu2019dlg, zhao2020idlg}.  
  
% On the other hand, text data presents several additional challenges \cite{tag} and we identify three key differences between text and image recovery settings: the much larger space of tokens possible at each position, concentration of sensitive information in specific tokens, and human tolerance to errors in pixel values as opposed to words.

% The techniques employed for text recovery range from early attempts using modified continuous optimization-based recovery to approaches that combine discrete and continuous optimization or genetic algorithms, and more recent techniques that exploit the low-rank nature of attention gradients to identify tokens at specific positions using a span check \cite{tag}, \cite{balunovic2022lamp}, \cite{petrov2024dager}.

Text data, on the other hand, presents additional challenges, such as a larger token space, sensitive information concentrated in a few specific tokens, and low error tolerance in words as opposed to pixels \cite{tag}. Addressing the challenges, several techniques have been developed for text recovery \cite{tag, feng2024grab, balunovic2022lamp, petrov2024dager}. However, these attacks assume an honest-but-curious server that observes each client's individual gradient, and they degrade as the batch size grows. This makes these methods infeasible for client-side recovery, where gradients are aggregated over many clients, hence effectively making it a much larger batch of text to recover. 

The strongest recent method, DAGER \citep{petrov2024dager}, recovers text up to 128 sentences but hits a strict \emph{wall} once the number of tokens in the observed batch approaches the model's hidden dimension $d$. Moreover, in the client-side setting, with \textsc{FedAvg} \cite{mcmahan2017fedavg}, even with modest participation, the number of tokens $C$ in the aggregate over clients greatly exceeds $d$. Our aim is to enable recovery from an aggregate of gradients and make client-side recovery possible.

   % \anote{I'd also suggest using some math notation with batch size x seq\_len x num\_clients here, its fine? If you think otherwise, phrase it well}

% We identify two factors contributing to this limitation brought on by the batch size. First, as the number of tokens ($b \cdot s \cdot n$) increases, the gradient directions associated with individual tokens become linearly dependent, making individual tokens difficult to separate from the aggregate. Second, frequent tokens (articles, prepositions, subword fragments) appear many times within and across the sentences of a batch, so gradient matching alone cannot determine which occurrence of the token maps to which position. We learn from these observations to design a novel technique aimed at overcoming the aggregation bottleneck and enable text recovery in a client-side attack model.
% \anote{"increasingly fused" - can we rewrite this sentence? a reviewer is gonna be confused af by that. "multiplicity of common tokens" - you can keep this phrase, but i think you need to define it clearly here.}

% \textbf{In this work}, we propose a novel gradient inversion attack specifically for text recovery at larger batch sizes and demonstrate, for the first time, a client-side honest-but-curious attack model for text recovery. \anote{"for the first time" -> make it something like "as far as we are aware, we are the first attack that [blah blah blah

Our key insight is to develop a pipeline in which, once some sentences are precisely recovered, their gradients are subtracted from the aggregate, leaving a cleaner signal for recovering the rest. Our contributions are as follows.

\begin{itemize}[nosep, leftmargin=*]
    \item We formulate a gradient inversion attack specifically designed for \emph{above-the-wall} text recovery ($C > d$), enabling an honest-but-curious client adversary to reconstruct other clients' text from aggregated model updates.
    % \item We investigate why text recovery breaks down at high batch sizes, and through ablations identify specific causes behind the aggregation bottleneck such as the multiplicity problem.  not needed
     \item We also introduce a novel attack pipeline that iteratively applies this insight, refining the residual gradient over successive rounds to recover more text from the aggregate.
    \item We demonstrate through an extensive evaluation that a single client adversary in an FL setup can exactly recover private training sentences from aggregate gradients, with 100\% precision. Our method is the first to succeed above the \textit{wall}, where existing attacks recover nothing exactly, and it outperforms them below it as well. 
% We evaluate our method on parallel-residual models, including the Pythia and GPT-NeoX families, along with more general architectures such as GPT-2 and Llama 2 in our ablation studies. We evaluate our method across the CoLA, Rotten Tomatoes, and SST-2 datasets  last three lines not needed
\end{itemize}
% \textbf{In this work}, 
% % Many realistic fine-tuning scenarios involve smaller-sized models, and our technique works especially well for the parallel-residual class of models that are common among open small-scale models \anote{I'd avoid this sentence, it's not adding anything here.}. 

% . \anote{This should be a sentence in the key contributions.}

% \textbf{Key Contributions} In this work we make the following key contributions: 

% \anote{Overall, this section is too long, but it's fine for now. Do not shorten it, I'll work on shortening it later on. Put all the information you think is relevant here for now.}

\section{Related Work}
In this section, we discuss existing attacks for text recovery using gradients, followed by a discussion on how the number of tokens in the observed gradient limits these attacks.

\subsection{Data recovery from gradients}
Gradient inversion attacks were first introduced by DLG \cite{zhu2019dlg}. The idea behind this attack is to recover data from gradients by initializing random values for the data and optimizing them to match the observed gradients. Improvements to this technique were later proposed by analytically identifying the label \cite{zhao2020idlg} and improving gradient matching \cite{geiping2020inverting}. However, these methods are designed primarily for image recovery. As these techniques are optimization-based, their application to text is much more challenging than their application to images, which are represented in a continuous space. The discrete nature of text makes this a harder optimization problem, which limits the effectiveness of the above techniques \cite{petrov2024dager}. 

This limitation has motivated a growing body of research specifically focused on text recovery from gradients. 

TAG \cite{tag} was the first to specifically formulate a gradient inversion attempt for LLMs and uses the same optimization-based technique with a new regularization term, which prioritizes matching gradients in layers closer to the input. FILM \cite{gupta2022recovering} takes a different approach from optimization-based methods by first recovering a bag of words from the observed gradients and then using LM-guided beam search to reconstruct the original sentences. LAMP \cite{balunovic2022lamp} and GRAB \cite{feng2024grab} use an alternating continuous-discrete optimization strategy, using an LM-prior score and beam search over recovered tokens, respectively, for the discrete optimization part. DAGER \cite{petrov2024dager} exploits the low-rank nature of self-attention gradients to recover tokens at each position using a span check. A recent method, SOMP \citep{somp}, casts recovery from aggregated gradients as greedy sparse signal recovery: it employs LM-guided beam search and greedily selects the few candidate gradients that best explain the observed aggregate. Like the methods above, however, it remains a server-side attack.

\begin{figure*}[t]
    \centering
    \resizebox{\textwidth}{!}{%
        \input{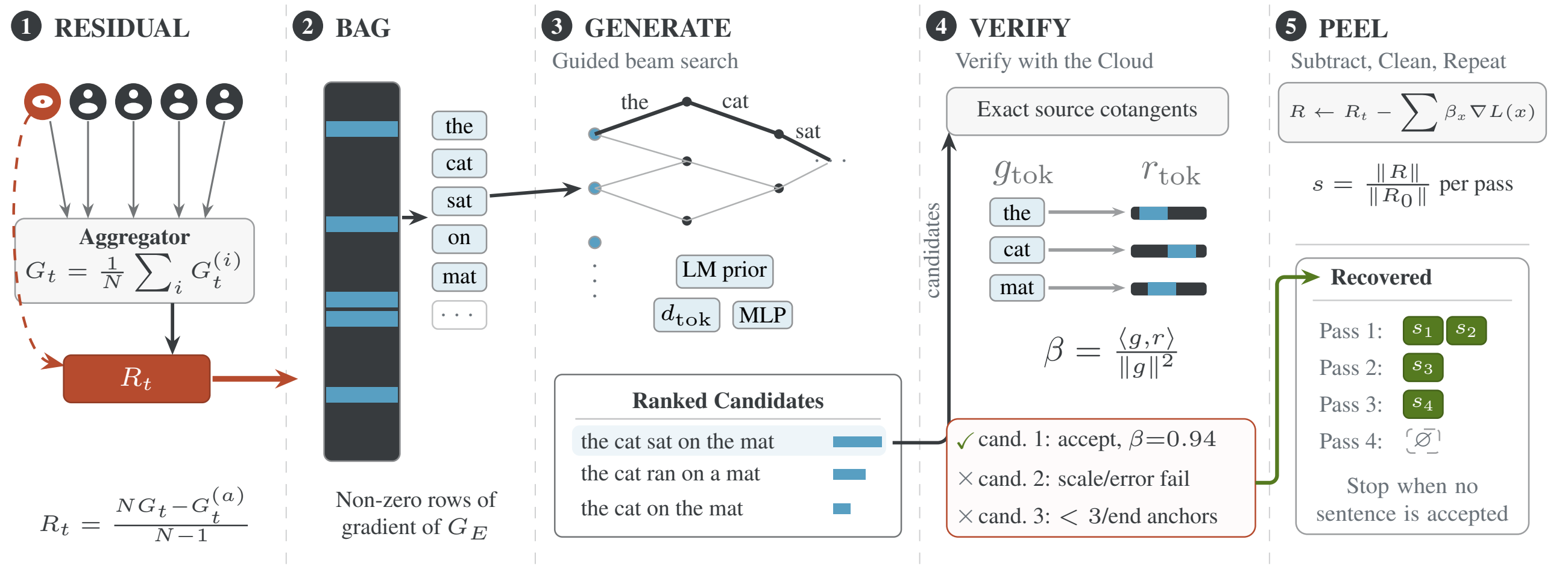}
    }
    \caption{\textbf{Overview of our method.} Step 1 outlines the method to isolate client gradients, while Step 2 extracts the bag of tokens (\S\ref{sub:bagoftoken}). Step 3 applies the candidate generation algorithm (\S\ref{sub:candgen}) to produce quality-ranked candidate strings, which we then verify with the gradient (\S\ref{subsec:verification}) in Step 4. Finally, Step 5 describes peeling (\S\ref{sec:peel}), the process which iteratively refines the residual gradient from the recovered sentences.}
    \label{fig:overview}
\end{figure*}

\subsection{Text recovery for large batch sizes}
Despite these developments, we observe that text recovery from gradients is handicapped by the size of the training batch. Older techniques such as TAG~\cite{tag} and FILM~\cite{gupta2022recovering} fail to recover meaningfully as the batch size becomes larger. DAGER~\cite{petrov2024dager}, which reports this degradation in its comparisons, achieves a near-exact reconstruction for much larger batches by checking whether candidate token embeddings lie in the span of the first-layer gradients, but this relies on the gradient being low-rank. Because the gradient's rank is bounded by the number of tokens, the check stops discriminating once that number reaches $d$. So far, this has been the \textit{wall} for text recovery.

\looseness=-1
In the client-side setting, the observed residual aggregates gradients from the tokens of every participating client, placing realistic \textsc{FedAvg} configurations well above the $d$ threshold. Consistent with this, existing baselines perform very poorly in our above-the-wall configurations and do not recover a single exact sentence, as explained in Section~\ref{sec:main} and Appendix~\ref{app:eval-metrics}. As stated earlier, the primary reason behind the failure of existing baselines is the noise introduced by aggregation within the gradient, which makes the content of individual tokens indistinguishable from the aggregate. As a result, with our method, we show that removing the gradient contribution of already-recovered sentences allows further recovery of sentences that were previously hidden. This clearly indicates that it is the degree of aggregation that limits recovery, and lowering it helps us recover more sentences.
\section{Methodology}

\subsection{Threat Model}
\label{sec:threat-model}
We consider an honest-but-curious client $\mathrm{adv}$ participating in federated
full fine-tuning of a parallel-residual LLM under the \textsc{FedAvg} protocol
\cite{mcmahan2017fedavg}. In each round $t$, the server and clients hold model
parameters $\theta_t$. Every participating client $j$ computes its raw gradient
$G_t^{(j)}$ on its local data $X_t^{(j)}$ at parameters $\theta_t$, and the server aggregates and
broadcasts the model update $\theta_{t+1}$, as given in Eq.~\eqref{eq:fedavg}:
\begin{equation}
\theta_{t+1} \;=\; \theta_t - \eta\, G_t,
\qquad
G_t \;=\; \frac{1}{c} \sum_{j=1}^{c} G_t^{(j)}
\label{eq:fedavg}
\end{equation}
where $c$ is the number of participating clients and $\eta$ is the learning rate. The update $G_t$ is standard \textsc{FedAvg} aggregation, i.e.\ a plain
average of raw client gradients for the loss function $\mathcal{L}$, and the attacker $\mathrm{adv}$ contributes its own
term $G_t^{(\mathrm{adv})}$. Because both $\theta_t$ and $\theta_{t+1}$ are broadcast to every client, the adversary derives
$G_t = (\theta_t - \theta_{t+1})/\eta$
without any additional assumptions on server behavior, and recovers the residual $R_t$, as given in Eq.~\eqref{eq:residual}:
\begin{equation}
R_t \;=\; \frac{c\, G_t - G_t^{(\mathrm{adv})}}{c-1}
     \;=\; \frac{1}{c-1} \sum_{j \neq \mathrm{adv}} G_t^{(j)}
\label{eq:residual}
\end{equation}
\looseness=-1
the average of the gradients of the other $c-1$ clients. The
adversary's goal is to reconstruct the other clients' data
$X_t=\{X_t^{(j)} : j \neq \mathrm{adv}\}$
from $R_t$ and model parameters $\theta_t$. The parameters $\theta_t$ are public, so the
weights, architecture, and tokenizer are available to the attacker.
These are the observations available to any participant in a
standard federated fine-tuning deployment.
Our methodology depicts a single round, so we omit the round index $t$ in what follows, writing $\theta = \theta_t$ for the parameters and $R_0 = R_t$ for the residual.

\begin{algorithm}[t]
\caption{Top-level recovery}
\label{alg:total}
\textbf{Input:} $\theta$ -- parameters broadcast in the attacked round;
$R_0$ -- residual gradient, the average of the other clients.\\
\textbf{Output:} $\hat{X}$ -- recovered sentences, each with a scale $\beta_x$.\\[0.6em]
1: $\hat{X} \leftarrow \emptyset$\\
2: $R \leftarrow R_0$\\
\textbf{repeat}\\
3: \quad $\mathcal{C} \leftarrow \textsc{Generate}(R,\theta)$  \\
4: \quad $\mathcal{A} \leftarrow \textsc{Verify}(R,\mathcal{C})$ \\
5: \quad \textbf{if} $\hat{X}\cup\mathcal{A}=\hat{X}$ \textbf{then return} $\hat{X}$\\
6: \quad $\hat{X} \leftarrow \hat{X}\cup\mathcal{A}$\\
7: \quad \textbf{peeling:} $R \leftarrow R_0 - \sum_{(x,\beta_x)\in\hat{X}}\beta_x\nabla\mathcal{L}(x;\theta)$\\
\textbf{forever}
\end{algorithm}

\subsection{Overview}
Given the residual \(R_0\) and public model parameters \(\theta\) established in the threat model, our approach recovers sentences through an iterative loop composed of candidate generation, gradient-based verification of candidates, and denoising of gradients through peeling. Figure~\ref{fig:overview} illustrates a pass of the recovery loop, which is further formalized in Algorithm~\ref{alg:total}.

At each iteration, \(\textsc{Generate}\) identifies a bag of tokens $B$ and proposes candidate sentences from the current residual using a beam search. Then,  \(\textsc{Verify}\) identifies candidates consistent with the observed gradients.
%and estimates a scale factor \(\beta_x\) for each accepted sentence, indicating the magnitude of its contribution to the aggregate gradient. 
Newly identified sentences are added to the recovered pool, and 
in the peeling step, we recompute the working residual by subtracting the estimated gradient contributions of all sentences recovered so far from the original residual \(R_0\), giving us a cleaner residual to recover additional sentences. 
%This removes the gradient contributions already explained by the recovered sentences, allowing subsequent iterations to focus on recovering sentences whose contributions remain in the residual. 
 When an iteration yields no new sentences, the process terminates. 

We customize this attack pipeline for the parallel-residual model class, but we also evaluate its applicability to sequential-residual models in Appendix~\ref{app:non-parallel-evals}. The following sections detail each of its components, starting with identification of the bag of tokens.

\subsection{Bag of Tokens Extraction}
\label{sub:bagoftoken}
The combined corpus of tokens that all the clients trained on is termed the bag of tokens. To identify the bag of tokens, in the case of untied models, where the input embedding $W_{\text{in}}$ and unembedding $W_{\text{out}}$ matrices are distinct (see Appendix \ref{app:ext-rw}), we just select the non-zero rows of the input gradient matrix, which reveals tokens that were involved in training \cite{melis2019exploiting, gupta2022recovering}. We provide a formalism for untied models in Appendix \ref{app:bagoftokens}.

The bag of tokens reveals which tokens appeared in the client data, but not the context they appeared in. To reconstruct sentences, we therefore compute a token quantity, called the \textit{token direction}, which captures information on the tokens that followed each token in the client data. We define this quantity in Lemma \ref{lem:tokdir}.

\begin{lemma}[Token directions]
\label{lem:tokdir}
In a parallel-residual architecture, we can use the first MLP layer gradients to recover, for every token $\mathrm{tok}$ in the bag $\mathcal{B}$, using a least-squares formulation, $d_{\mathrm{tok}}$ where
\[
d_{\mathrm{tok}} \;=\; \sum_{\text{occurrences of } \mathrm{tok} } \frac{\partial \mathcal{L}}{\partial h^{(1)}},
\]
where $h^{(1)}$ is the output of the first transformer block at that occurrence of $\mathrm{tok}$ in the input. $d_{\mathrm{tok}}$ captures information about the words that followed $\mathrm{tok}$ by aggregating the updates to the hidden states at positions where $\mathrm{tok}$ appeared as input.
The proof and the least squares formulation are provided in Appendix~\ref{app:tokdir}.

\end{lemma}
The solve for the \(d_{\mathrm{tok}}\) values is done simply on this bag. 

With a tied embedding, the problem of identification of the bag of tokens is slightly more difficult. As the model uses the same embedding and unembedding matrix (i.e., $W_{\text{in}} = W_{\text{out}}$), it has two gradient paths. Since the LM-head softmax contributes gradients to every row (forcing Lemma \ref{lem:tokdir} to be invalid), the embedding gradient is no longer sparse, making it difficult to distinguish tokens that appeared in the training data from those that did not. We therefore identify the bag of tokens jointly with their $d_{\mathrm{tok}}$ values by performing a least-squares solve over the top-$\mathbb{K}$ rows with the largest gradient magnitudes. Tokens that did not appear as inputs come back with \(d_{\mathrm{tok}}\) near zero and are dropped. The rest of the tokens form the bag of tokens. We can expect a high recall from this method, but since we consider the top-$\mathbb{K}$ rows, we do not expect a very high precision. We verify this in Appendix~\ref{app:non-parallel-evals}.

Having identified the bag of tokens, we next construct a pool of candidate sentences $\mathcal{C}$, aiming to include as many of the private client sentences as possible.

\subsection{Candidate Generation}
\label{sub:candgen}

% We are not concerned with precision here, since verification is done later. 
For generating the \textit{pool} of candidate sentences $\mathcal{C}$, we use beam search on the bag of tokens, guided by language-model (LM) fluency and the unembedding gradients.

Beam search is a search technique that explores the space of sentences by maintaining a set, or \emph{beam} of $w$ prefixes, extending each prefix by every possible next token at every step, ranking all the extensions by a score, and keeping the top $w$ to form the next \emph{beam} until a maximum length terminates the search.

Let $\sigma(\hat{x}, \mathrm{tok})$ be the score for extending a prefix $\hat{x}$ with a token $\mathrm{tok}$. We rank a partial sentence by the sum of the scores used to build it. The score is built by combining two signals, as shown in Eq.~\eqref{eq:score}. The first is a cosine similarity between the final hidden state $h_{\hat{x}}$ produced by the partial sentence $\hat{x}$ of the beam, and $o_{\mathrm{tok}}$, the row for token $\mathrm{tok}$ in the residual for the unembedding matrix $W_{\mathrm{out}}$, both passed through a whitening transformation $\phi(\cdot)$.  The cosine similarity lets us gauge whether $\mathrm{tok}$ follows $\hat{x}$, as Lemma~\ref{lem:cloud} shows. The second is plain LM fluency.
\begin{equation}
\sigma_\lambda(\hat{x},\mathrm{tok}) \;=\;
\underbrace{\big\langle \phi(h_{\hat{x}}),\, \phi(-o_{\mathrm{tok}}) \big\rangle}_{\text{cosine similarity}}
\;+\; \lambda\,
\underbrace{\log p_\theta(\mathrm{tok} \mid \hat{x})}_{\text{language-model fluency}}.
\label{eq:score}
\end{equation}

\begin{lemma}[Unembedding matrix rows record the hidden states of prefixes preceding each token]
\label{lem:cloud}
For every token $\mathrm{tok}$, the aggregated gradient residual
satisfies\footnote{With tied embeddings, the observed row also contains
the input-embedding gradient of $\mathrm{tok}$; see
Appendix~\ref{app:cloud}.}
\[
-o_{\mathrm{tok}}
=
\underbrace{\sum_{\forall x \text{ s.t. } y_x = \mathrm{tok}} \alpha_x h_x}
_{\text{ prefixes followed by } \mathrm{tok}}
-
\underbrace{\sum_{\forall x} \alpha_x
p_\theta(\mathrm{tok}\mid x)\,h_x}
_{\text{background over all prefixes}},
\]
where $o_{\mathrm{tok}}$ is the row for $\mathrm{tok}$ in the residual for $W_{\mathrm{out}}$, $y_x$ is the token that follows prefix $x$ in a training sentence, $h_x$ is its final hidden state, and $\alpha_x>0$ are the aggregation weights. In words, $-o_{\mathrm{tok}}$ is the sum of the hidden states of the prefixes that $\mathrm{tok}$ followed, minus a background term shared by all rows.
The proof is provided in Appendix~\ref{app:cloud}.
\end{lemma}

%where $i$ runs over every token position in the other clients' data,
%$s_{i+1}$ is the token that follows position $i$, $h_i$ is the final
%hidden state at position $i$ (the vector the output head reads), and
%$\alpha_i > 0$ is a constant fixed by the aggregation.

The first term of the lemma is what our score uses. When a partial sentence $\hat{x}$ in the beam matches a prefix $x$ of a sentence in client data, since the model is causal, the hidden state $h_{\hat{x}}$ we compute after reading $\hat{x}$ is exactly the hidden state $h_x$ the client produced at that position. Since $h_x$ is one of the vectors summed in $-o_{\mathrm{tok}}$, how closely $h_{\hat{x}}$ points along $-o_{\mathrm{tok}}$ is evidence that $\mathrm{tok}$ follows $\hat{x}$.

\subparagraph{Whitening Transformation $\phi(\cdot)$.} Due to the second term of Lemma~\ref{lem:cloud} present in every row, comparing $o_{\mathrm{tok}}$ with $h_{\hat{x}}$ directly through cosine similarity fails to give a clear signal. We therefore apply a whitening transformation $\phi(\cdot)$ to $h_{\hat{x}}$ and $-o_{\mathrm{tok}}$, which rescales each dimension, centers the rows, and projects out their dominant shared directions, following \citet{mu2018allbutthetop}. This removes the component common to all rows of the gradient. We discuss this transformation in more detail in Appendix~\ref{app:whitening}.

\paragraph{Language-model fluency $\lambda \log p_\theta(\mathrm{tok} \mid \hat{x})$.}
The cosine term is only a token-level statistical signal, and sometimes scores incorrect continuations on par with correct ones. LM fluency helps tell the two apart, so we add it to the score with a small weight $\lambda$.

In every round of the beam search, we take the top $w$ ways, ranked by the score $\sigma_\lambda$ in Eq.~\eqref{eq:score}, to extend the beam using tokens from the bag, and the top $w$ ways to end it with sentence-final tokens. The first becomes the beam for the next round, and the second is added to the pool $\mathcal{C}$. At the beam's maximum length, every remaining prefix is ended and added to $\mathcal{C}$. For exact implementation and formal details, we refer the readers to Appendix~\ref{app:beam}. 

We also notice that only a small fraction of sentences in $\mathcal{C}$ are actually client sentences. We next begin to verify which candidates were part of the training data.

\subsection{Verification}
\label{subsec:verification}

Verification rests on the insight that, if a sentence appeared in the training set, all of its gradient contributions should be present in the residual with a consistent scaling factor.

\looseness = -1
To verify a candidate $x = (s_1, \dots, s_n)$, we compute the gradients from training on it, where at each position $i < n$ the model predicts the next token $s_{i+1}$. We keep two gradients at each position $i$. One is $\partial \mathcal{L}/\partial e_i$, the gradient with respect to the input embedding $e_i$ fed at position $i$, i.e., the row of $W_{\mathrm{in}}$ for token $s_i$, taken separately at each position where it appears. The other is $\partial \mathcal{L}/\partial h^{(1)}_i$, the gradient with respect to the output of the first transformer block $h^{(1)}_i$ at position $i$.

However, the residual quantities we compare these gradients against are recorded per token rather than per position, so the comparison must also be done per token.

We call each distinct token in $s_1, \dots, s_{n-1}$ an \emph{anchor}. The last token $s_n$ is excluded since it is only ever predicted, never used to predict, and so receives no input gradient. Each anchor is intended to provide independent evidence that the candidate sentence appeared in training. For an anchor $\mathrm{tok}$, we sum each of the two gradients over every position $i$ where $s_i = \mathrm{tok}$, giving us two per-token quantities, $g_{\mathrm{tok}}$ and $u_{\mathrm{tok}}$,
\begin{align*}
  g_{\mathrm{tok}} &= \sum_{i \,:\, s_i = \mathrm{tok}} \frac{\partial\mathcal{L}}{\partial e_i}, &
  u_{\mathrm{tok}} &= \sum_{i \,:\, s_i = \mathrm{tok}} \frac{\partial\mathcal{L}}{\partial h^{(1)}_i}.
\end{align*}
The residual holds the same two sums, taken over every position where $\mathrm{tok}$ was input across all sentences in the aggregate. The first is $r_{\mathrm{tok}}$, row $\mathrm{tok}$ of the residual for $W_{\mathrm{in}}$. The second is the token direction $d_{\mathrm{tok}}$ (Lemma~\ref{lem:tokdir}), the sum of gradients with respect to the first-block hidden states at those positions.

\begin{figure}[!ht]
    \resizebox{\columnwidth}{!}{%
\begin{tikzpicture}[x=1cm,y=1cm]
\definecolor{vermillion}{HTML}{B64B2E}
\definecolor{blugre}{HTML}{557A20}
\definecolor{ink}{HTML}{373B3E}
\definecolor{sky}{HTML}{589FC2}
\definecolor{grey}{HTML}{697178}

\renewcommand{\familydefault}{\sfdefault}
\newcommand{\sxs}{\fontsize{6}{7.2}\selectfont}
\newcommand{\tns}{\fontsize{5}{6}\selectfont}
\newcommand{\dL}[1]{\partial\mathcal{L}\!/\!\partial #1}

\tikzset{
  >={Stealth[length=2pt,width=1.8pt]},
  lab/.style={text=grey, font=\tns, anchor=east, inner sep=0pt, align=right},
  tgt/.style={text=grey, font=\sxs, inner sep=0.6pt},
  hid/.style={draw=vermillion!70, fill=vermillion!10, rounded corners=1.2pt,
              minimum width=0.80cm, minimum height=0.36cm, inner xsep=0pt, inner ysep=1pt,
              text=vermillion!80!black, font=\sxs},
  inp/.style={draw=ink!55, fill=sky!18, rounded corners=1.2pt,
              minimum width=0.80cm, minimum height=0.22cm, inner sep=0pt,
              text=ink, font=\sxs},
  anc/.style={draw=blugre!80, fill=blugre!10, rounded corners=1.4pt,
              inner xsep=1.2pt, inner ysep=1.4pt, align=center,
              text=blugre!60!black, font=\tns},
  wire/.style={->, draw=ink!60, line width=0.35pt},
  feed/.style={->, draw=blugre!80, line width=0.4pt, rounded corners=1.5pt},
}

% ---- layout knobs ----
\def\dx{1.12}                                   % column pitch
\def\xL{0.44}                                   % right edge of row labels
\def\yT{1.62} \def\yH{1.20} \def\yI{0.80}       % target / hidden / input rows
\def\yA{0.52}                                   % top of single-occurrence anchors

\node[lab] at (\xL,\yT) {target};
\node[lab] at (\xL,\yH) {hidden};
\node[lab] at (\xL,\yI) {input};

% one column per position i: input token s_i -> hidden h^{(1)}_i -> predicts next token
\foreach \tok/\nxt/\i in {the/cat/1, cat/sat/2, sat/on/3, on/the/4, the/mat/5}{
  \pgfmathsetmacro{\x}{\xL+0.48+(\i-1)*\dx}
  \node[tgt] (t\i) at (\x,\yT) {\nxt};
  \node[hid] (h\i) at (\x,\yH) {$h^{(1)}_{\i}$};
  \node[inp] (e\i) at (\x,\yI) {\tok};
  \draw[wire] (h\i.north) -- (t\i.south);
  \draw[wire] (e\i.north) -- (h\i.south);
}

% anchors that occur once: g and u are a single term
\foreach \tok/\i in {cat/2, sat/3, on/4}{
  \node[anc, anchor=north] (a\i) at (e\i |- 0,\yA)
    {$g_{\text{\tok}}\,{=}$\\[0.6pt]$\dL{e_{\i}}$\\[2pt]
     $u_{\text{\tok}}\,{=}$\\[0.6pt]$\dL{h^{(1)}_{\i}}$};
  \draw[feed] (e\i.south) -- (a\i.north);
}
\node[lab] at (\xL,0 |- a3.center) {anchor};

% anchor that occurs twice: g and u sum over both positions
\node[anc, anchor=north] (athe) at ([yshift=-0.14cm]a3.south)
  {$g_{\text{the}}=\dL{e_1}+\dL{e_5}$\\[1pt]
   $u_{\text{the}}=\dL{h^{(1)}_1}+\dL{h^{(1)}_5}$};
\draw[feed] (e1.south) |- (athe.west);
\draw[feed] (e5.south) |- (athe.east);

\end{tikzpicture}%
}
    \caption{Calculation of $g_{\mathrm{tok}}$ and $u_{\mathrm{tok}}$ by summing the input-embedding and hidden-state gradients over all positions containing the token $\mathrm{tok}$.}
    \label{fig:verify}
\end{figure}
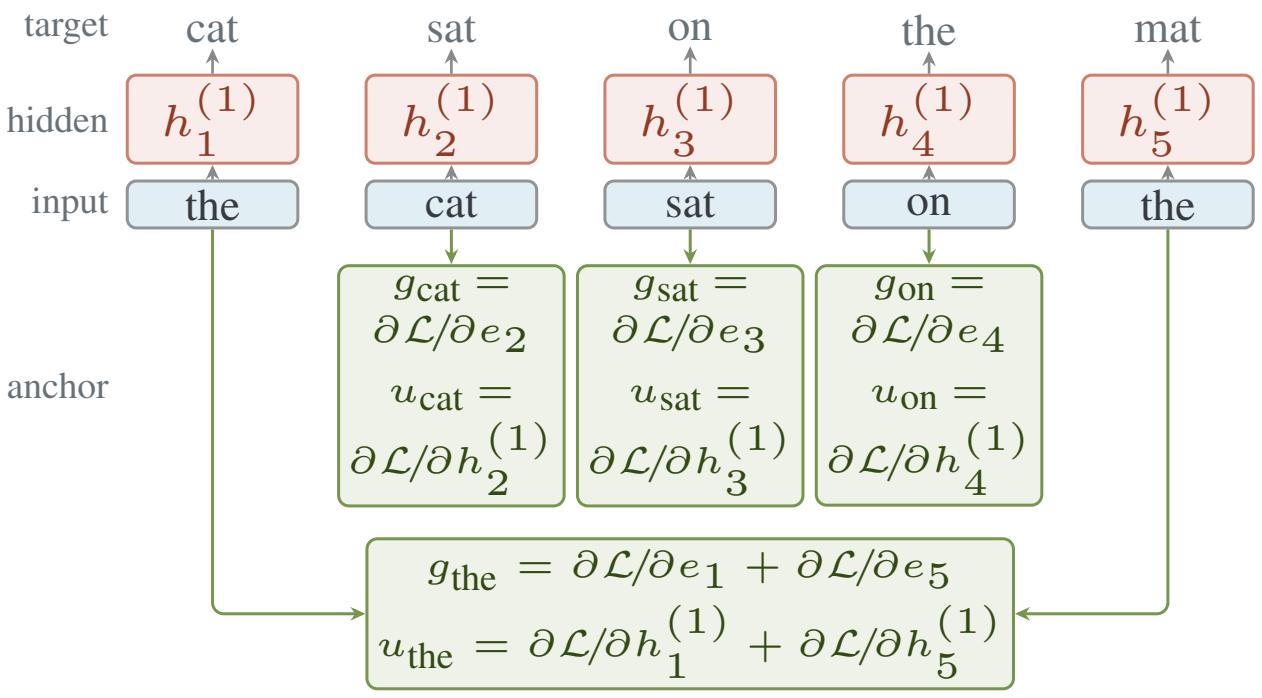

As shown in Figure~\ref{fig:verify}, in the sentence ``the cat sat on the mat'', \emph{the} appears at positions 1 and 5, so
$g_{\mathrm{the}} = \partial\mathcal{L}/\partial e_1 + \partial\mathcal{L}/\partial e_5$ and $u_{\mathrm{the}} = \partial\mathcal{L}/\partial h^{(1)}_1 + \partial\mathcal{L}/\partial h^{(1)}_5$.

If the candidate sentence was present in training, both residual quantities should match their candidate counterparts under the same scaling factor $\beta$, i.e., $r_{\mathrm{tok}} \approx \beta\, g_{\mathrm{tok}}$ and $d_{\mathrm{tok}} \approx \beta\, u_{\mathrm{tok}}$.

We first estimate this scale $\beta_{\mathrm{tok}}$ separately for each anchor by fitting $g_{\mathrm{tok}}$ to $r_{\mathrm{tok}}$ with least squares,
\[
\beta_{\mathrm{tok}}=\frac{\langle g_{\mathrm{tok}}, r_{\mathrm{tok}}\rangle}{\|g_{\mathrm{tok}}\|^2}.
\]
We then check that the same scale also fits the other pair, i.e., $d_{\mathrm{tok}} \approx \beta_{\mathrm{tok}}\, u_{\mathrm{tok}}$.

\looseness = -1
We retain anchors whose gradient pairs match under a positive scaling factor $\beta_{\mathrm{tok}}$, subject to error thresholds, and evaluate the candidate as a whole using its surviving anchors.

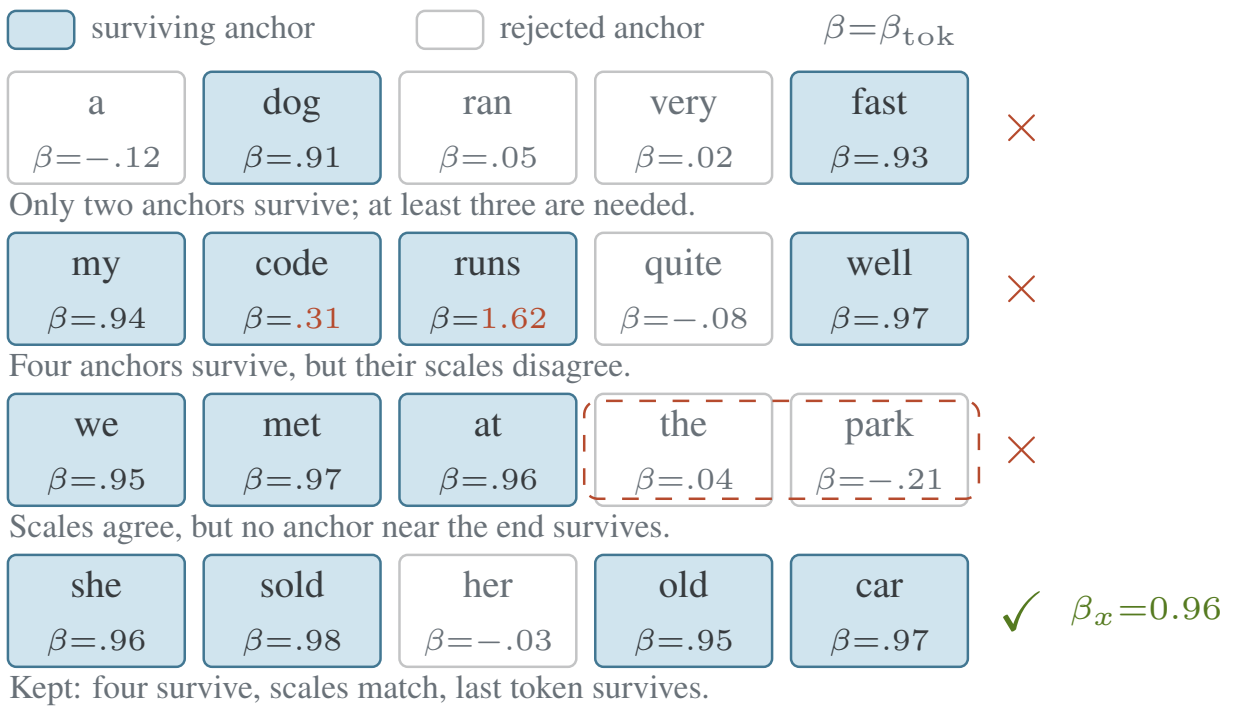
\begin{figure}[!htbp]
    \centering
    \resizebox{\columnwidth}{!}{%
        % \documentclass[tikz,border=1pt]{standalone}
% \usepackage[T1]{fontenc}
% \usepackage{lmodern}
% \usepackage{amsmath,amssymb}
% \usepackage{xcolor,graphicx}
% \usetikzlibrary{arrows.meta}
% \begin{document}

\begin{tikzpicture}[x=1cm,y=1cm]
\definecolor{vermillion}{HTML}{B64B2E}
\definecolor{blugre}{HTML}{557A20}
\definecolor{ink}{HTML}{373B3E}
\definecolor{sky}{HTML}{589FC2}
\definecolor{grey}{HTML}{697178}

\renewcommand{\familydefault}{\sfdefault}
\newcommand{\wrd}{\fontsize{7}{8}\selectfont}
\newcommand{\bet}{\fontsize{5.5}{6.5}\selectfont}
\newcommand{\nte}{\fontsize{6}{7.2}\selectfont}

\tikzset{
  % surviving anchor: blue box; rejected anchor: plain box
  keep/.style={draw=sky!75!black, fill=sky!28, text=ink},
  drop/.style={draw=ink!30, fill=white, text=grey},
  tokbox/.style={rounded corners=1.5pt, line width=0.45pt,
                 minimum width=1.12cm, minimum height=0.52cm,
                 inner xsep=0pt, inner ysep = 2pt, align=center},
  note/.style={text=grey, font=\nte, anchor=north west, inner sep=0pt},
  bad/.style={text=vermillion},
}

% one token box: \tokb{x}{y}{style}{word}{beta}
\newcommand{\tokb}[5]{%
  \node[tokbox,#3] at (#1,#2)
    {{\wrd\strut #4}\\[-2.5pt]{\bet $\beta{=}{#5}$}};}

\def\dx{1.24}    % box pitch
\def\xr{6.42}    % x of the verdict mark
\def\dy{1.02}    % row pitch

% ---- legend ----
\node[tokbox,keep,minimum width=0.42cm,minimum height=0.24cm] at (0.21,0.62) {};
\node[note,anchor=west] at (0.52,0.62) {surviving anchor};
\node[tokbox,drop,minimum width=0.42cm,minimum height=0.24cm] at (2.80,0.62) {};
\node[note,anchor=west] at (3.11,0.62) {rejected anchor};
\node[note,anchor=west] at (5.15,0.62) {$\beta{=}\beta_{\mathrm{tok}}$};

% ---- row 1: too few survivors ----
\def\y{0}
\foreach \w/\b/\s [count=\i from 0] in {a/{-.12}/drop, dog/.91/keep, ran/.05/drop,
                                        very/.02/drop, fast/.93/keep}{
  \tokb{0.56+\i*\dx}{\y}{\s}{\w}{\b}}
\node[text=vermillion, font=\fontsize{10}{10}\selectfont] at (\xr,\y) {$\times$};
\node[note] at (0,\y-0.4) {Only two anchors survive; at least three are needed.};

% ---- row 2: scales disagree ----
\def\y{-\dy}
\tokb{0.56+0*\dx}{\y}{keep}{my}{.94}
\tokb{0.56+1*\dx}{\y}{keep}{code}{\textcolor{vermillion}{.31}}
\tokb{0.56+2*\dx}{\y}{keep}{runs}{\textcolor{vermillion}{1.62}}
\tokb{0.56+3*\dx}{\y}{drop}{quite}{-.08}
\tokb{0.56+4*\dx}{\y}{keep}{well}{.97}
\node[text=vermillion, font=\fontsize{10}{10}\selectfont] at (\xr,\y) {$\times$};
\node[note] at (0,\y-0.4) {Four anchors survive, but their scales disagree.};

% ---- row 3: no survivor near the end ----
\def\y{-2*\dy}
\foreach \w/\b/\s [count=\i from 0] in {we/.95/keep, met/.97/keep, at/.96/keep,
                                        the/.04/drop, park/{-.21}/drop}{
  \tokb{0.56+\i*\dx}{\y}{\s}{\w}{\b}}
\draw[vermillion, dashed, line width=0.45pt, rounded corners=2.5pt]
  (0.56+3*\dx-0.63,\y-0.31) rectangle (0.56+4*\dx+0.63,\y+0.31);
\node[text=vermillion, font=\fontsize{10}{10}\selectfont] at (\xr,\y) {$\times$};
\node[note] at (0,\y-0.4) {Scales agree, but no anchor near the end survives.};

% ---- row 4: kept ----
\def\y{-3*\dy}
\foreach \w/\b/\s [count=\i from 0] in {she/.96/keep, sold/.98/keep, her/{-.03}/drop,
                                        old/.95/keep, car/.97/keep}{
  \tokb{0.56+\i*\dx}{\y}{\s}{\w}{\b}}
\node[text=blugre, font=\fontsize{10}{10}\selectfont] at (\xr,\y) {$\checkmark$};
\node[note] at (0,\y-0.4) {Kept: four survive, scales match, last token survives.};
\node[text=blugre, font=\nte, anchor=west] at (\xr+0.18,\y) {$\beta_x{=}0.96$};

\end{tikzpicture}%

% \end{document}
    }
    \caption{Candidate filtering in \textsc{Verify}. Blue boxes denote surviving anchors; each box shows its $\beta_{\mathrm{tok}}$ value.}
    \label{fig:candidates}
\end{figure}

We consider a candidate verified only when (i)~at least $p$ of its anchors survive (we set $p=3$ in our implementation), (ii)~their scales lie close to one another, and (iii)~one of them sits near the end of the candidate, which ensures that the candidate is a complete sentence rather than a prefix of one. Past works (Appendix~\ref{sec:poor-recovery}) failed to reproduce an entire sentence successfully, which is why we include the third condition above. Figure~\ref{fig:candidates} demonstrates this using examples.

For a successfully verified sentence, we refit a single scale $\beta_x$ over all its surviving anchors. \textsc{Verify} then returns the set of verified candidates along with their scales $\beta_x$, which indicate how strongly each sentence's gradient is present in the residual. These scales are used in the next step, peeling (Section~\ref{sec:peel}), to subtract each verified sentence's contribution from the residual and deliver a cleaner signal for the next iteration.

\subsection{Peeling}
\label{sec:peel}
Peeling is the mechanism that enables recovery of sentences beyond the initial set. The idea of peeling comes from our observation that in high-batch-size regimes, gradients from all the different sentences overlap with each other, and this leads to noise in the gradients, hampering recovery.\par
Using the set of sentences \(\hat{X}\) recovered so far, we subtract their gradient contributions from the residual \(R_0\). The next working residual can hence be defined as $R$ where,
% \textbf{The Math.} Line 7 in Algorithm 1 describes peeling.
\[
R = R_0 - \sum_{(x,\beta_x)\in\hat{X}} \beta_x \nabla\mathcal{L}(x;\theta).
\]
$\nabla\mathcal{L}(x;\theta)$ represents the gradient of the next-token loss due to sentence $x$, summed over all its positions. Multiplying this by $\beta_x$ obtained from the verification step allows us to remove the gradient due to sentence $x$ by the amount that is actually present in $R_0$. This weighted sum is then done over all the sentences recovered so far and subtracted from the original residual $R_0$ to obtain the peeled gradient $R$ for the next iteration to work with.

The gradients of recovered sentences in $\hat{X}$ dominate the residual, causing subsequent recovery rounds to repeatedly identify the same sentences. Peeling subtracts their estimated gradient contributions from $R_0$, exposing weaker signals from previously unrecovered sentences.

\textbf{Cleaning the embedding rows.} After the subtraction, a tiny positive leftover might remain in the embedding row of a token. If the leftover norm is tiny compared to the original,
\[
\|R[W_{\mathrm{in}}]_{\mathrm{tok}}\| < \tau_{\mathrm{c}} \,\|R_0[W_{\mathrm{in}}]_{\mathrm{tok}}\|,
\]
the embedding row is zeroed out, so that the particular token is not considered as a seed in the next iteration.

\paragraph{Summary.} 
To summarize our entire pipeline, starting from the residual gradient, we identify the bag of
tokens (\S\ref{sub:bagoftoken}) and use gradient-guided beam search to generate candidate sentences
(\S\ref{sub:candgen}). We then verify candidates by checking whether their gradient contributions match the
residual under a consistent scale (\S\ref{subsec:verification}). Peeling (\S\ref{sec:peel}) subtracts the estimated contributions of verified sentences, exposing weaker signals for further recovery. This loop continues on the same model update until no new sentences are verified, as shown in Algorithm~\ref{alg:total}. Appendix~\ref{app:qualitative}
(Figure~\ref{fig:qualitative-leak}) presents qualitative
examples of text recovered by this pipeline.

\section{Evaluations}
\label{sec:experiments}

\begin{figure*}[!t]
    \centering
    \includegraphics[width=\linewidth]{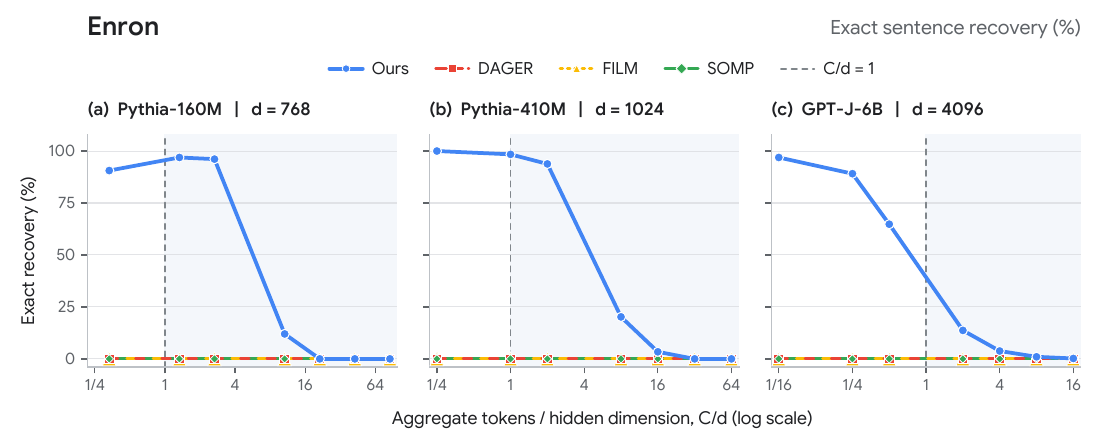}
    \caption{Recovery rates (\%) for three models on the Enron dataset. The x-axis is $C/d$, with $C=b \cdot c \cdot s$ aggregated tokens ($b$: batch size per client, $c$: clients, $s$: sequence length) and model hidden dimension $d$. We also report the BLEU-2 and ROUGE-2 scores and compare them against the specified baselines in Figures \ref{fig:bleu2_enron} and \ref{fig:rouge2_enron} respectively.}
    \label{fig:recovery_enron}
\end{figure*}

\begin{figure*}[!b]
    \centering
    \includegraphics[width=\linewidth]{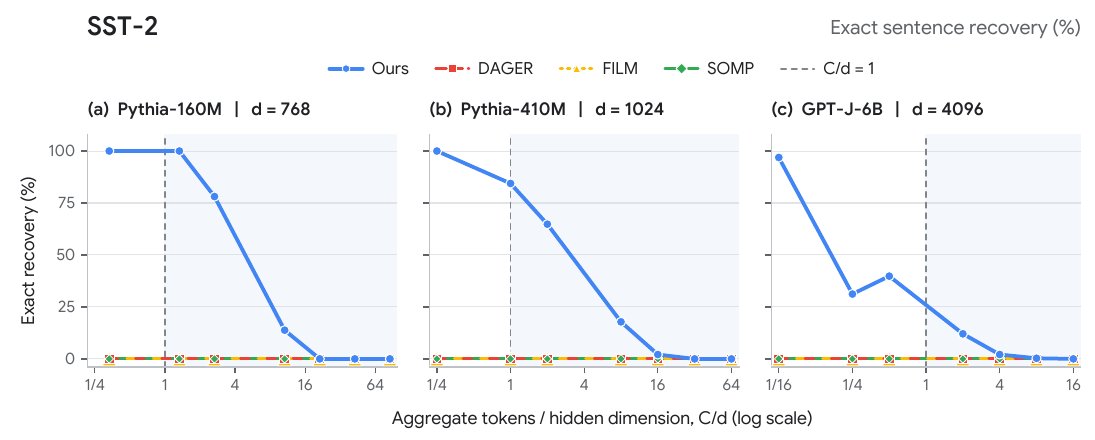}
    \caption{Recovery rates (\%) for three models on the SST-2 dataset. The x-axis is $C/d$, with $C=b \cdot c \cdot s$ aggregated tokens ($b$: batch size per client, $c$: clients, $s$: sequence length) and model hidden dimension $d$. We also report the BLEU-2 and ROUGE-2 scores and compare them against the specified baselines in Figures \ref{fig:bleu2_sst2} and \ref{fig:rouge2_sst2} respectively.}
    \label{fig:recovery_sst2}
\end{figure*}

\looseness=-1
We evaluate our method for recovering private training sentences under the threat model described in Section \ref{sec:threat-model}. We compare recovery rates and precision against existing attacks both below and above the wall, and study how recovery scales with the number of clients, batch size, and sequence length of sentences. We further conduct ablation studies to assess the contribution of individual components and examine the effectiveness of existing defenses against our attack.

\subsection{Experimental setup}
\label{sec:setup}
\paragraph{Models and data.}

\looseness=-1
We evaluate our method on \texttt{Pythia-160M} \cite{biderman2023pythia}, \texttt{Pythia-410M} \cite{biderman2023pythia}, and \texttt{GPT-J-6B} \cite{gpt-j}, with hidden dimensions $d=768$, $1024$, and $4096$, respectively. Since the primary focus of this work is on parallel-residual architectures, we use the above models. We also present some results for non-parallel architectures in Appendix \ref{app:non-parallel-evals}. Our evaluation covers several private datasets, including SST-2 \cite{socher-etal-2013-recursive} movie reviews, Enron email bodies \cite{klimt2004enron}, and PubMed medical text \cite{cohan-etal-2018-discourse}, covering sentiment analysis, email correspondence, and biomedical writing.

\paragraph{Configurations.} We denote each configuration by $(c,b,s)$, where $c$ is the number of clients contributing to the observed residual, $b$ is the per-client batch size, and $s$ is the sequence length. Table \ref{tab:experimental_configs} lists the configurations and their relation to each model's hidden dimension. These settings range from 256 to 65,536 tokens, allowing us to evaluate recovery both below the wall and under aggregation substantially beyond it. We also describe all our default hyperparameters and other configuration settings in Appendix \ref{app:other-configs}.

\paragraph{Evaluation metrics.}
We report exact token-level sentence recovery to measure the accuracy and coverage of the recovered tokens. However, exact sentence recovery does not capture partially correct reconstructions. We therefore also compare BLEU-2 and ROUGE-2 to quantify the overlap between each method's outputs and the original client sentences. Appendix~\ref{app:eval-metrics} defines these metrics formally.

\paragraph{Baselines.}
We compare against DAGER \cite{petrov2024dager},
FILM \cite{gupta2022recovering}, and SOMP \cite{somp}.
Although these methods were developed for server-side recovery and face limitations as the number of aggregated tokens grows, we apply them in this situation to assess whether they can recover any text in this setting, including above the hidden-dimension wall. We describe how we run these baselines in Appendix~\ref{app:baseline-info}.

\subsection{Recovery of private texts}
\label{sec:main}

\begin{figure*}[!t]
    \centering
    \includegraphics[width=\linewidth]{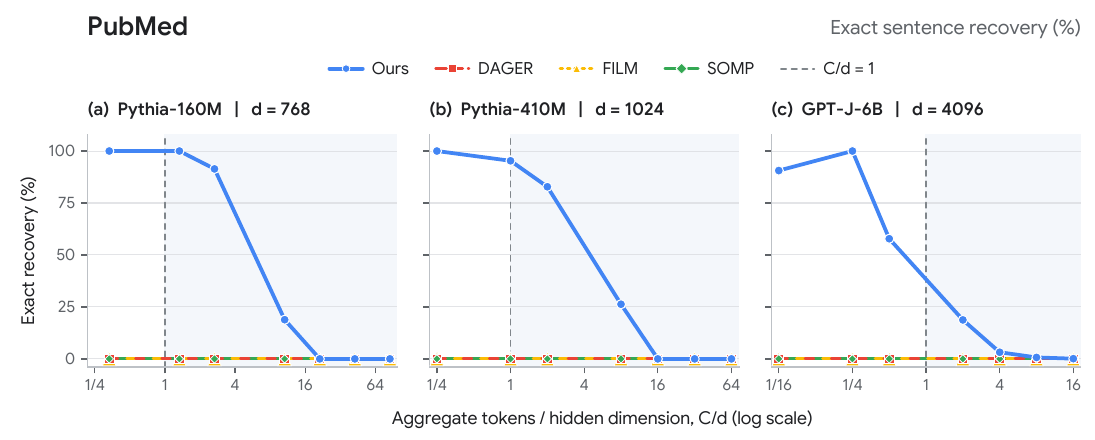}
    \caption{Recovery rates (\%) for three models on the PubMed dataset. The x-axis is $C/d$, with $C=b \cdot c \cdot s$ aggregated tokens ($b$: batch size per client, $c$: clients, $s$: sequence length) and model hidden dimension $d$. We also report the BLEU-2 and ROUGE-2 scores and compare them against the specified baselines in Figures \ref{fig:bleu2_pm} and \ref{fig:rouge2_pm}, respectively.}

    \label{fig:recovery_pubmed}
\end{figure*}

\begin{figure*}[!b]
    \centering
    \includegraphics[width=\textwidth]{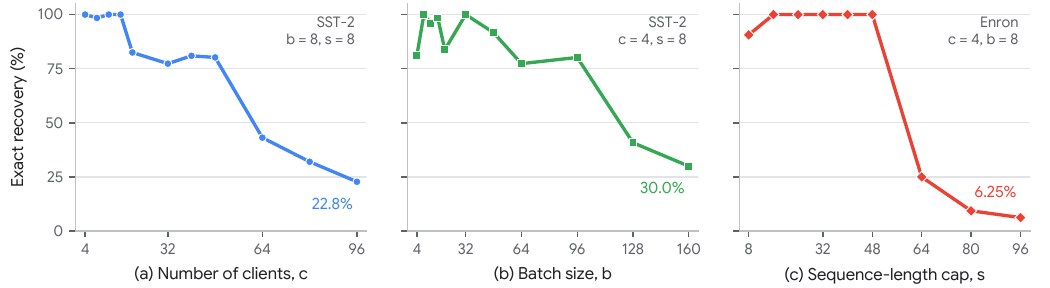}
    \caption{Exact sentence recovery for \texttt{Pythia-160M}, varying one aggregation parameter at a time. (a) Clients $c$ on SST-2, with $b=8$ and $s=8$. (b) Batch size $b$ on SST-2, with $c=4$ and $s=8$. (c) Sequence-length cap $s$ on Enron, with $c=4$ and $b=8$. Lines connect the measured recovery rates.}
    \label{fig:scaling-cbs}
\end{figure*}

Figures \ref{fig:recovery_enron}, \ref{fig:recovery_sst2},
and \ref{fig:recovery_pubmed} show recovery results on Enron, SST-2, and PubMed, respectively. 
We plot the number of exactly recovered sentences against
$C/d$, where $C=b \cdot c \cdot s$ is the total number of aggregated tokens within the gradient and $d$ is the model's hidden dimension. Configurations with $C/d>1$ lie above the \emph{wall}.

\looseness = -1
We observe that our method outperforms all existing baselines both below and above the wall. DAGER, FILM, and SOMP do not produce any precisely recovered sentences in any evaluated configuration, even below the wall. To compare with partial recovery, our method also achieves higher BLEU-2 and ROUGE-2 in almost all settings (Figures~\ref{fig:bleu2_enron}--\ref{fig:rouge2_pm}). We explain these failures of the baselines seen in the plots in Appendix~\ref{sec:poor-recovery}.

Although our method enables exact recovery above the wall, its
effectiveness still declines under sufficiently large
aggregations. As $C/d$ increases further, the number of
recovered sentences approaches zero. Thus, our method extends the regime in which recovery is possible, but remains limited when the aggregate contains too many tokens relative to the model's hidden dimension. We discuss the reasons for this decline in Appendix \ref{app:moving-the-wall}.

\subsection{Scaling clients, batch size, and sequence length}
\label{subsec:scaling-cbs}
We vary the number of clients $c$, batch size per client $b$, and sequence-length cap (tokens) $s$ separately, holding the other two fixed (Figure~\ref{fig:scaling-cbs}). All experiments use \texttt{Pythia-160M} \cite{biderman2023pythia}, and the client and batch sweeps use SST-2; the length sweep uses Enron because its longer texts allow us to evaluate larger caps.

The client and batch sweeps yield similar recovery at matched $M=c \cdot b$, which is representative of the number of sentences in the gradient ($43.2\%$ versus $40.8\%$ at $M=512$), suggesting that aggregate sentence count captures much of the difficulty in this setting. Neither sweep produces false accepts, so increasing the aggregate size reduces recall while preserving verification precision.

On the sequence length sweep on Enron, recovery remains $100\%$ for the tested caps $s=16$ to $48$, then drops significantly at $s=96$. Precision also falls from $80\%$ to $66.7\%$ at $s=64 \rightarrow 96$, respectively, showing that longer-sequence settings impair both reconstruction and verification.

\subsection{Ablation studies}
In this section, we rigorously evaluate every step of our recovery pipeline to validate if it was necessary for recovery or not. We also discuss any other interesting findings from it. Unless stated otherwise, we use \texttt{Pythia-160M} on SST-2 with $(c,b,s)=(4,32,16)$, which recovers 100 sentences (78.1\%) with sentence-level precision $1.00$. We also do further detailed experiments on hyperparameter selection and other sensitivity studies in Appendix \ref{app:hyperparam-sensitivities}. We also do further efficiency and cost analysis studies in Appendix \ref{sec:efficiency}.

\paragraph{Bag of words.} 
For a model with untied embeddings such as \texttt{Pythia-160M}, bag-of-tokens extraction identifies 866 of 921 reference tokens, achieving precision $1.00$ and recall $0.94$. On the other hand, we also validate that this method works for tied models such as GPT-2 as well. We capture $875$ of the $906$ reference tokens (recall $0.97$), but at low precision ($0.21$) because we capture a large top-$\mathbb{K}$ (4096) that exceeds the true token set.

\begin{figure}
    \centering
    \includegraphics[width=\linewidth]{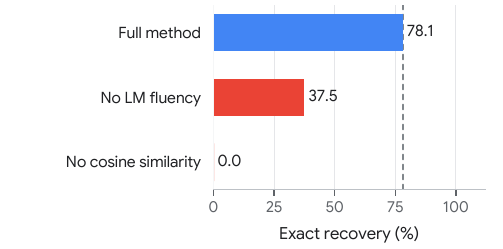}
    \caption{Ablation studies on the components of candidate generation \textsc{Generate}.}
    \label{fig:abl-candid}
\end{figure}

\paragraph{Scoring in candidate generation.}
\looseness = -1
Figure \ref{fig:abl-candid} evaluates each scoring component by individually removing it. Removing the cosine similarity reduces exact recovery from 100 sentences to zero. Removing LM fluency reduces recovery to 48 sentences. We also study the effect of whitening in Appendix \ref{app:whitening-strange}.

\begin{figure}
    \centering
    \includegraphics[width=\linewidth]{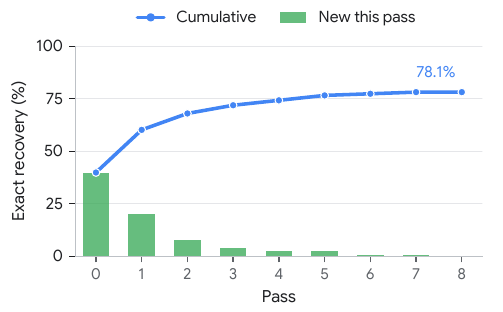}
    \caption{Peeling and verification in our pipeline increase our recovery rate by $1.96\times$, making it an especially critical part of our method.}
    \label{fig:abl-peel}
\end{figure}

\paragraph{Verification and peeling.}
Iterative recovery via peeling yields $1.96\times$ as many exact sentences as the initial pass alone. The initial pass (after \textsc{Verify}) recovers 51 sentences. Peeling enables seven further productive passes, recovering another 49 sentences and bringing the total to 100 (Figure \ref{fig:abl-peel}), making peeling a critical part of the pipeline.

\subsection{Defenses and discussions}
\label{subsec:defense}

We examine whether common gradient perturbations and clipping mechanisms can mitigate our attack, and assess its sensitivity to local training configurations.

\begin{figure}
    \centering
    \includegraphics[width=\linewidth]{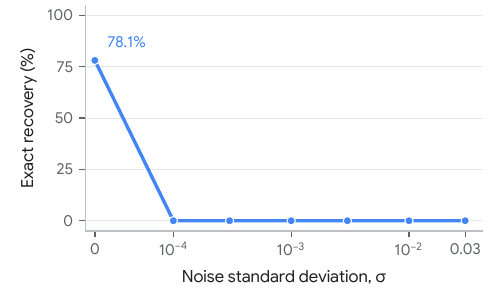}
    \caption{Effect of gradient perturbation on exact recovery. Adding Gaussian noise to shared gradients reduces exact recovery to zero across all tested nonzero noise levels $\sigma$.}
    \label{fig:defense}
\end{figure}

\paragraph{Perturbation and clipping.}
Inspired by differential privacy \cite{Mironov_2017}, for \texttt{Pythia-160M} and SST-2 with the configuration $(4, 32, 16)$, adding Gaussian noise $\epsilon\sim\mathcal{N}(0,\sigma^2 I)$ to the shared gradient reduces exact recovery from $100/128$ ($78.1\%$) to zero at every tested nonzero noise level (Figure \ref{fig:defense}).

\paragraph{Clipping and local updates.}
Gradient clipping \cite{koloskova2023revisitinggradientclippingstochastic} is also a popular defense, but we notice that recovery remains $100/128$ and is unaffected by clipping. This shows that this method is robust to certain preliminary FL defenses as well. 

\begin{figure}
    \centering
    \includegraphics[width=\linewidth]{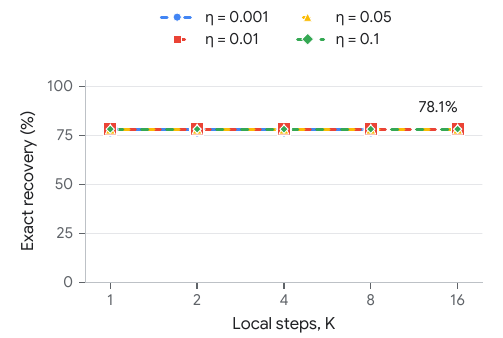}
    \caption{Robustness of recovery performance across 20 \textsc{FedAvg} configurations with varying local-step counts and learning rates.}
    \label{fig:def-agg}
\end{figure}

\paragraph{Optimization algorithms \& sensitivity to local updates.}
\looseness = -1
Apart from \textsc{FedAvg}, we also compare our recovery rates against other popular FL aggregation algorithms, such as \textsc{FedSGD}, in Appendix \ref{app:aggregation-algos}. Moreover, we also notice that recovery is unchanged across the 20 reported \textsc{FedAvg} configurations, covering diverse local-step counts and learning rates, as seen in Figure \ref{fig:def-agg}. Thus, neither aggregate rescaling nor these local-update settings prevent recovery in our method.

% \begin{table}[h]
% \centering
% \caption{Attack wall-clock time (seconds) per configuration.}
% \label{tab:timing}
% \small
% \setlength{\tabcolsep}{6pt}
% \begin{tabular}{@{}lrrrr@{}}
% \toprule
% Model & $32\!:\!8$ & $64\!:\!16$ & $128\!:\!16$ & $512\!:\!16$ \\
% \midrule
% Pythia-160M & 39  & 882   & 2{,}150  & 3{,}474 \\
% Pythia-410M & 56  & 1{,}019 & 2{,}111 & 8{,}279 \\
% GPT-J-6B    & 268 & 7{,}644 & 22{,}148 & 11{,}853 \\
% \bottomrule
% \end{tabular}
% \end{table}

\section{Conclusions \& Limitations}

\paragraph{Limitations and Broad Impact.} Our main limitation is the model architecture on which our attack is fully effective, namely the parallel-residual class. A reduced variant also recovers text from sequential-residual models, but with lower precision (Appendix~\ref{app:non-parallel-evals}). Extending the full pipeline to a broader range of architectures remains future work. We also discuss other limitations regarding recovery in Appendix \ref{app:moving-the-wall}.

However, what we introduce, for awareness purposes, is a framework that recovers sentences from a heavy aggregation by gradually denoising the gradients. This framework could support further attacks that exploit properties of other architectures, extending the privacy risks posed by participating clients. Aggregation alone should therefore not be treated as a privacy guarantee in federated learning.

\paragraph{Conclusions.} This paper demonstrates, for the first time, exact recovery of private training data from an aggregate of gradients of multiple clients by an honest-but-curious client. Our results show that keeping training data local and aggregating client updates can still permit exact text recovery by a participating client. By verifying candidate sentences and peeling their gradient contributions from the aggregate, our method exposes private text even beyond the hidden-dimension barrier that limits existing attacks. These findings highlight the need for explicit privacy protections in federated fine-tuning.

% Acknowledgements should only appear in the accepted version.
% \section*{Acknowledgements}

% In the unusual situation where you want a paper to appear in the
% references without citing it in the main text, use \nocite
\newpage
\bibliography{references}

\bibliographystyle{mlsys2025}

\section*{AI Usage Disclosure}
% We use AI for assistance in editing this paper (only grammar and shi), also with minor help in deploying our code on GPUs and debugging code. 

% Write this section up.

We used LLMs and AI-based coding agents to debug our code and to run minor tests for our experiments. Their contributions included revising code, troubleshooting implementation issues, and assisting with deployment and execution on GPU infrastructure. For manuscript
preparation, we used LLMs for minor language editing, including grammar correction, rephrasing, and improvements to clarity.

The authors remain responsible for the correctness of the code, the validity and interpretation of the experimental results, and all statements and conclusions in the paper.

\clearpage
\appendix
\label{sec:appendix}

\section{Extended Related Works}
\label{app:ext-rw}
\paragraph{Optimization and discrete search:}  
Early works like TAG \cite{tag} focused only on continuously optimizing the candidate embeddings to match the target gradient. Since the process of tokenization followed by embedding is non-differentiable, only optimizing continuously might yield correct tokens, but their order cannot be determined. Methods like GRAB and FILM \cite{feng2024grab, gupta2022recovering} use discrete optimizations alternately and attempt to reorder the tokens found during continuous optimization. DAGER \cite{petrov2024dager} exploits the low-rank nature of gradient updates and uses an efficient subspace projection trick to quickly find the bag of words. It then use beam search guided by information from the next layers to piece together the bag to give coherent sentences. The second step is much more computationally expensive, and the first step (fast bag-of-words extraction) fails when the token budget exceeds the hidden dimension.

\paragraph{Tied and untied models:}
Language models differ in whether the input embedding matrix $W_{in}$ and the LM head unembedding matrix $W_{out}$ are separate parameters. This choice strongly affects what a client's gradient reveals about its text. In \emph{untied} models (e.g., Pythia), row $v$ of $\nabla_{W_{in}}\mathcal{L}$ receives a contribution only from positions $i$ where $s_i = v$, where $s_i$ indicate the token. The set of non-zero rows therefore equals the set of tokens in the client's input, which exposes the bag of words exactly, with no optimization and no dependence on the token budget or hidden dimension.

In \emph{tied} models (e.g., \texttt{GPT-2}, \texttt{Gemma}, \texttt{Llama-3.2-1B}), $W_{in} = W_{out}$, and the shared matrix also receives the gradient of the output softmax layer. Due to the final LM head softmax, the gradient rows are dense, i.e., almost all rows of the vocabulary are non-zero; the sparsity pattern that identifies input tokens in the untied case is lost. The input-side contribution is superimposed on a dense output-side term, and the two cannot be separated by inspecting the embedding gradient alone. Tied embeddings are standard in smaller and on-device models, which are the models most likely to be trained or fine-tuned directly on clients.

\begin{figure*}
    \centering
    \includegraphics[width=0.85\linewidth]{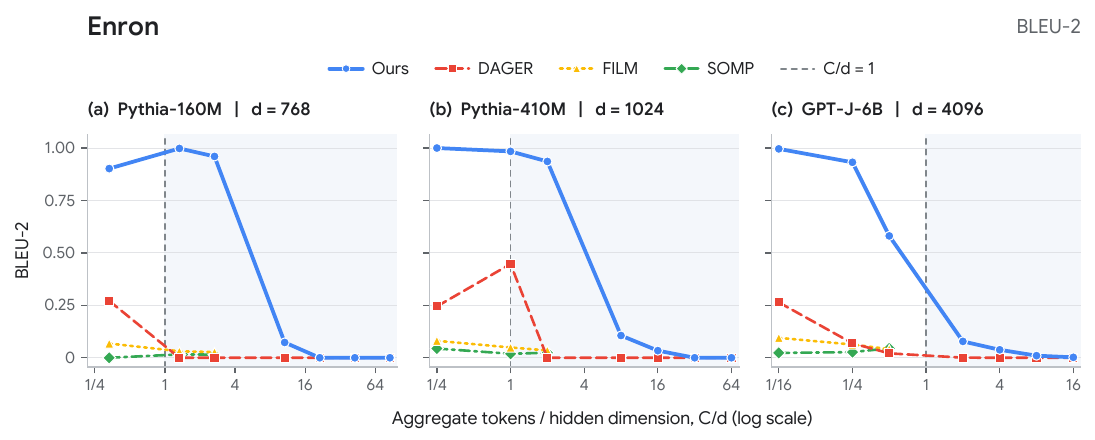}
    \caption{BLEU-2 for three models on the Enron dataset.}
    \label{fig:bleu2_enron}
\end{figure*}

\begin{figure*}
    \centering
    \includegraphics[width=0.85\linewidth]{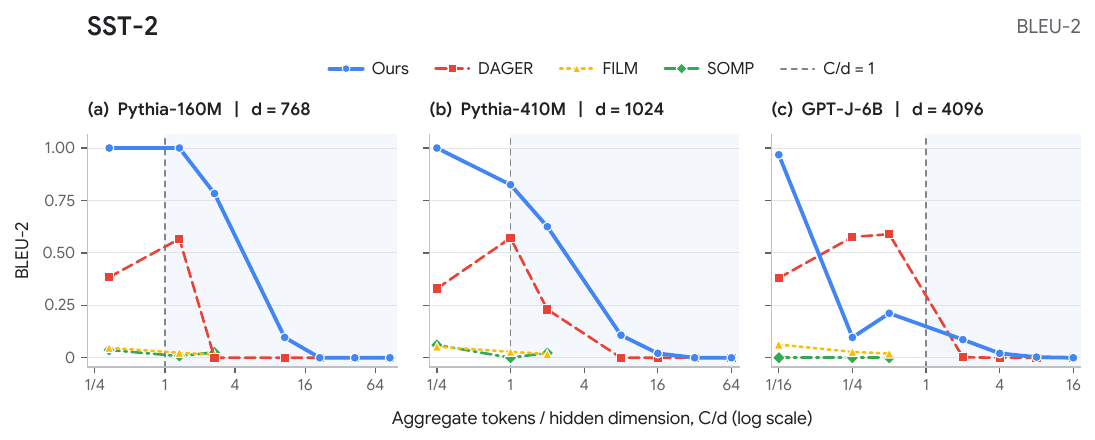}
    \caption{BLEU-2 for three models on the SST-2 dataset.}
    \label{fig:bleu2_sst2}
\end{figure*}

\begin{figure*}
    \centering
    \includegraphics[width=0.85\linewidth]{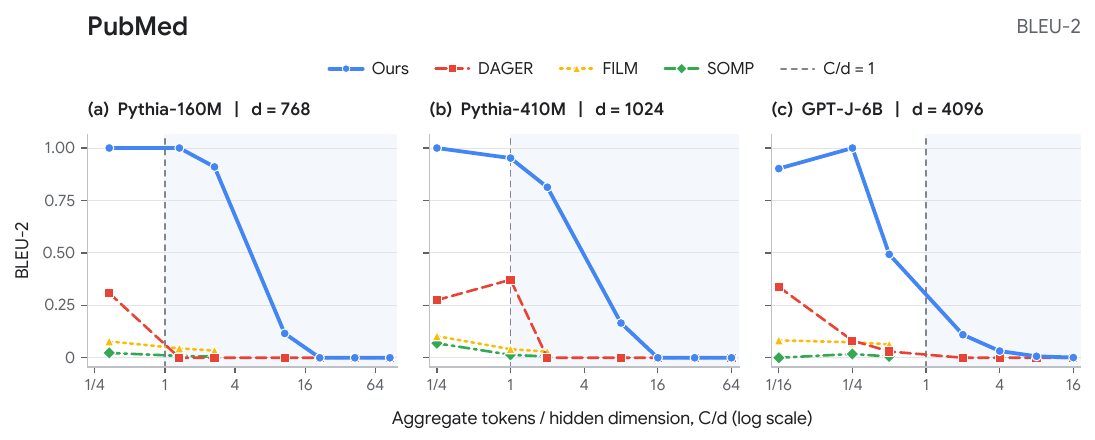}
    \caption{BLEU-2 for three models on the PubMed dataset.}
    \label{fig:bleu2_pm}
\end{figure*}

\begin{figure*}
    \centering
    \includegraphics[width=0.85\linewidth]{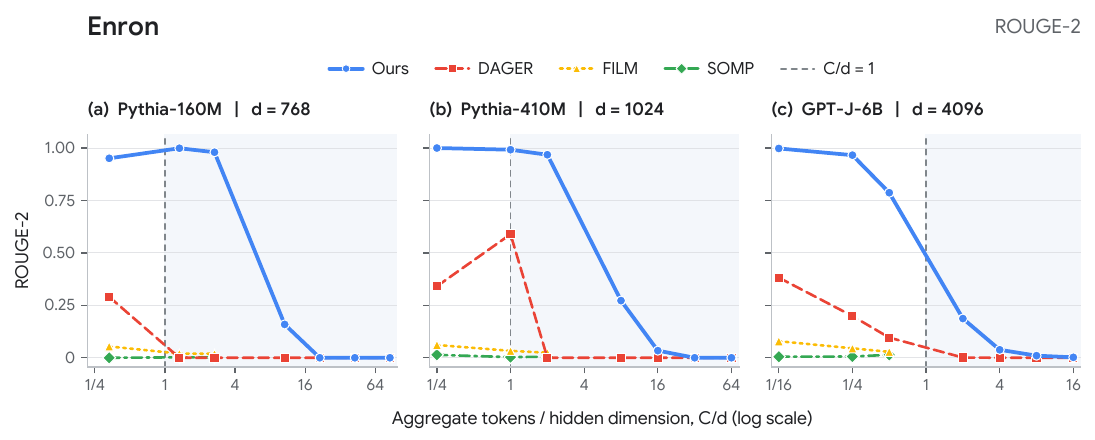}
    \caption{ROUGE-2 for three models on the Enron dataset.}
    \label{fig:rouge2_enron}
\end{figure*}

\begin{figure*}
    \centering
    \includegraphics[width=0.85\linewidth]{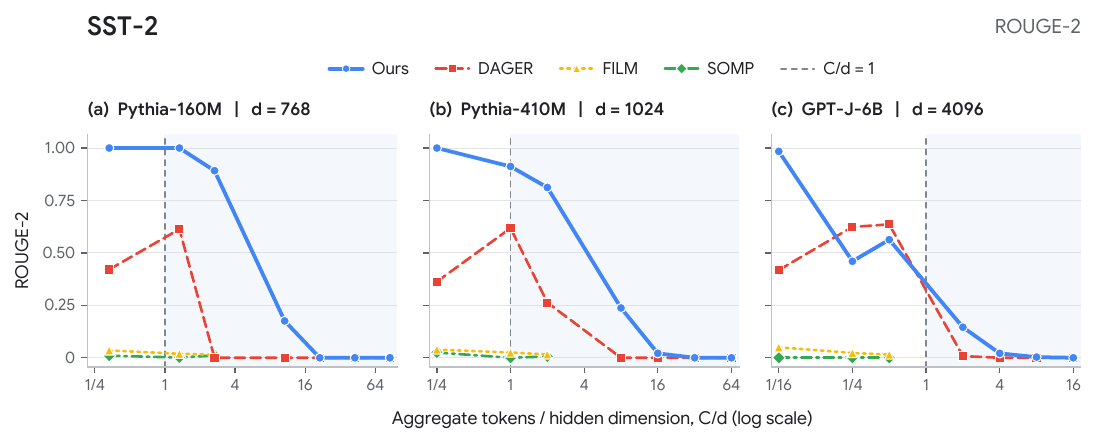}
    \caption{ROUGE-2 for three models on the SST-2 dataset.}
    \label{fig:rouge2_sst2}
\end{figure*}

\begin{figure*}
    \centering
    \includegraphics[width=0.85\linewidth]{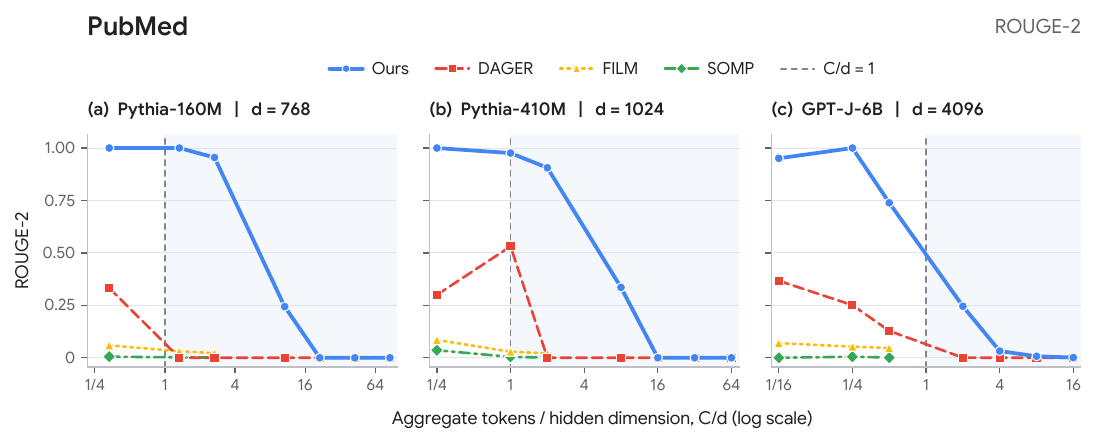}
    \caption{ROUGE-2 for three models on the PubMed dataset.}
    \label{fig:rouge2_pm}
\end{figure*}

\section{Beam Search Algorithm / Usage}
\label{app:beam}
Algorithm~\ref{alg:beam} shows how we implement the search. To start the
process, every token in the bag seeds its own beam of width $w$. At every step,
each partial sentence in the beam is extended by every token in the bag $\mathcal{B}$ to
form the set of extension candidates $\mathcal{Y}$. Of all these extensions, the
best $w$ continuations, ranked by the cumulative step score, form the next beam.
The best $w$ ways to end the sentence are committed to the pool $\mathcal{C}$,
ranked by their average step score. For ending a sentence during the beam
search, we only consider sentence-final punctuation tokens or end-only tokens
(defined below), provided the prefix has at least three tokens. Committing a
sentence does not remove its prefix from the beam. The beam search continues
until it reaches a maximum length $L$, at which point we allow any token to end
the sentence and commit it to the pool. Chains with a repeating three-token loop are penalized during the beam search.

Partial sentences are extended only with tokens from the bag $B$.
Sentence-final punctuation tokens are those whose text ends in \texttt{.},
\texttt{?}, or \texttt{!}. The set of end-only tokens holds the 256 tokens
outside $B$ with the largest rows in the residual for $W_{\mathrm{out}}$. These
are tokens whose gradients are high only in rows of $W_{\mathrm{out}}$ and not
$W_{\mathrm{in}}$, so they have only appeared in an output position, i.e.\ as
the last token of a sentence.

\paragraph{Pool formation.} After the search, duplicate sentences are removed
and the pool is ranked by average step score, from which we only keep the top
$N$ sentences. We find better results when we run the search for various values
of $\lambda$ in Eq.~\eqref{eq:score} and create the final pool as the union of
the pools of each individual run.

\begin{algorithm}[t]
\caption{\textsc{BeamSearch}$(\sigma)$}
\label{alg:beam}
\begin{algorithmic}[1]
\Require bag $B$; score $\sigma$; beam width $w$; maximum length $L$
\Ensure pool $\mathcal{C}$ of candidate sentences
\State $\mathcal{C} \leftarrow \emptyset$
\ForAll{tokens $s \in B$} \Comment{one beam per seed}
  \State beam $\mathcal{X} \leftarrow \{(s)\}$, with $q(s) = 0$
  \For{$\ell = 1$ \textbf{to} $L-1$}
    \State extend every $\hat{x} \in \mathcal{X}$ by every token $\mathrm{tok}$ into
      $\mathcal{Y} = \{\hat{x} \cdot \mathrm{tok}\}$ ($\hat{x}$ followed by $\mathrm{tok}$), with
      $q(\hat{x} \cdot \mathrm{tok}) = q(\hat{x}) + \sigma(\hat{x},\mathrm{tok})$
    \State $q(\hat{y}) \leftarrow q(\hat{y}) - 1$ for each $\hat{y} = \hat{x} \cdot \mathrm{tok} \in \mathcal{Y}$
      where the last three tokens of $\hat{x}$ repeat the three before them
    \State $\mathcal{X} \leftarrow \operatorname{top}_w(\mathcal{Y})$
    \If{$\ell \ge 3$ \textbf{or} $\ell = L-1$}
      \State $\mathcal{C} \leftarrow \mathcal{C} \cup
        \operatorname{top}_w\{\hat{y} \in \mathcal{Y} : \hat{y} \text{ ends a sentence}\}$,
        scored $q(\hat{y})/\ell$
    \EndIf
  \EndFor
\EndFor
\State \Return $\mathcal{C}$
\end{algorithmic}
\end{algorithm}

\section{Whitening Transformation}
\label{app:whitening}

% Move the full whitening derivation here:
% 1. Column-wise scaling
% 2. Mean removal
% 3. SVD and dominant-direction projection
% 4. Final transformation

We observe that the gradient rows are inherently susceptible to the low-rank geometry of the observed cloud. The averaged gradients \(R\) are dominated by high-variance noise that makes downstream
recovery nearly impossible. Before performing any comparisons, we pass all gradients through the denoising function $\phi$ that removes the top $k$ principal components contributing to the highest variance.        \par
   Let $O \in \mathbb{R}^{|\mathcal{T}| \times d}$ collect the residual rows
   $o_{\mathrm{tok}}$, and let $\nu_j$ be the standard deviation of column $j$ of $O$.
   We first rescale each head dimension to unit spread,                                                                                                            
   \begin{equation}    
   w_j \;=\; \frac{1}{\nu_j + \mu_\nu \cdot 10^{-3}},
   \qquad                         
   \tilde{o}_{\mathrm{tok}} \;=\; -\, o_{\mathrm{tok}} \circ w ,                                     
   \label{eq:colscale}                                              
   \end{equation}
   where $\mu_\nu$ is the median of $\{\nu_j\}$ and the small additive floor 
   protects near-constant dimensions.  After removing the mean
   $\bar{o} = \frac{1}{|\mathcal{T}|}\sum_{\mathrm{tok}} \tilde{o}_{\mathrm{tok}}$, an SVD of the centered                       
   cloud yields its principal directions                                                      
   $\tilde{O} - \mathbf{1}\bar{o}^{\top} = U \Sigma V^{\top}$; we discard the top $k$ directions,              \[                                             
   \Pi_k \;=\; V_k V_k^{\top},                                                                                                                                     
   \qquad                                                                                                                                                          
   V_k \;=\; \text{top-}k \text{ right singular vectors.}                                                        \]                                                     
   \label{eq:projector}                                                                                                                            
   The transform is then                                                                                                                                           
   \begin{equation}                                                                                                                                                
   \phi(v) \;=\;                                                                                                                                                   
   \frac{\big(v \circ w - \bar{o}\big) - \big(v \circ w - \bar{o}\big)\,\Pi_k}                                                                                     
        {\big\lVert \big(v \circ w - \bar{o}\big) - \big(v \circ w - \bar{o}\big)\,\Pi_k\big\rVert_2},                                                            
   \label{eq:phi}                                                                                                                                                  
   \end{equation}                                                                                                                                                  
    We term the entire pipeline of column rescaling, mean removal, projection out of the dominant subspace, and $\ell_2$ normalization as whitening. Since the rescaling is per-dimension, $\phi$ does not fully decorrelate the dimensions; we use the term in the spirit of \citet{ su2021whitening} and use it similar to \cite{mu2018allbutthetop}.

 \section{Proof for Token Directions}
\label{app:tokdir}

\noindent\textbf{Lemma 1 (Token directions).}
\textit{In a parallel-residual architecture, we can use the first MLP layer
gradients to recover, for every token $\mathrm{tok}$ in the bag $\mathcal{B}$,
using a least-squares formulation, $d_{\mathrm{tok}}$ where}
\[
  d_{\mathrm{tok}} = \sum_{\text{occurrences of tok}}
  \frac{\partial \mathcal{L}}{\partial h^{(1)}},
\]
\textit{where $h^{(1)}$ is the output of the first transformer block at that
occurrence of $\mathrm{tok}$ in the input.}

\begin{proof}
For the first transformer block, we assume:
\begin{enumerate}
    \item[1)] The model has a parallel-residual architecture, and positional
  information is applied only to the queries and keys inside attention
  (rotary embeddings). For an input sequence $(s_1, \dots, s_n)$ with input
  embeddings $e = (e_{s_1}, \dots, e_{s_n})$, the hidden state at position $i$ is
  \[
    h^{(1)}_i = e_{s_i} + \mathrm{Attn}\big(\mathrm{LN}_1(e)\big)_i
              + \mathrm{MLP}\big(\mathrm{LN}_2(e_{s_i})\big).
  \]
  \item[2)] Layer normalization acts on each token independently.
  \item[3)] No dropout is applied during the hidden state computation.
\end{enumerate}

By 1) and 2), the input to the MLP at position $i$, $\mathrm{LN}_2(e_{s_i})$,
depends only on its token $s_i$, and by 3) the activation $a_{s_i}$ of the
first MLP (the input to its output projection $W_{\mathrm{mlp},0}$) is a direct
function of its token id.

These activations can therefore be recomputed using the model weights.
Write $z_i = W_{\mathrm{mlp},0}a_{s_i}$ for the output projection at
position $i$. By 1), $z_i$ is added directly to the block output
$h^{(1)}_i$, so
\[
  \frac{\partial \mathcal{L}}{\partial z_i}
  = \frac{\partial \mathcal{L}}{\partial h^{(1)}_i}.
\]
Changing the entry $(W_{\mathrm{mlp},0})_{pq}$ changes only the
$p$-th coordinate of $z_i$, at rate $(a_{s_i})_q$. Since the same
projection is applied at every position, the chain rule gives
\[
  \frac{\partial \mathcal{L}}
       {\partial (W_{\mathrm{mlp},0})_{pq}}
  = \sum_i
    \frac{\partial \mathcal{L}}{\partial (z_i)_p}\,(a_{s_i})_q
  = \sum_i
    \frac{\partial \mathcal{L}}{\partial (h^{(1)}_i)_p}\,(a_{s_i})_q.
\]
Collecting these entries into a matrix yields
\begin{equation}
  G_{\mathrm{mlp},0}
  = \sum_i \frac{\partial \mathcal{L}}{\partial h^{(1)}_i}
    a_{s_i}^{\top}.
  \label{eq:position-sum}
\end{equation}

All occurrences of the same token share the same activation.
Grouping the terms by token therefore gives
\[
  G_{\mathrm{mlp},0}
  = \sum_{\mathrm{tok}\in\mathcal{B}}
    \underbrace{\left(
      \sum_{i:s_i=\mathrm{tok}}
      \frac{\partial \mathcal{L}}{\partial h^{(1)}_i}
    \right)}_{d_{\mathrm{tok}}}
    a_{\mathrm{tok}}^{\top},
\]
where the outer sum is over distinct tokens in the bag.
Since both $G_{\mathrm{mlp},0}$ and the activations are known, we solve
\[
  \{\widehat d_{\mathrm{tok}}\}
  \in \operatorname*{arg\,min}_{\{v_{\mathrm{tok}}\}}
  \left\|
    G_{\mathrm{mlp},0}
    - \sum_{\mathrm{tok}\in\mathcal{B}}
      v_{\mathrm{tok}}a_{\mathrm{tok}}^{\top}
  \right\|_F^2.
\]
The true directions $\{d_{\mathrm{tok}}\}$ give zero residual.
If the activations of the distinct tokens in $\mathcal{B}$ are
linearly independent, this solution is unique, so
$\widehat d_{\mathrm{tok}} = d_{\mathrm{tok}}$ for every token in the bag.
\end{proof}

\section{Proof for Output Rows}
\label{app:cloud}
\noindent\textbf{Lemma 2 (Un-embedding matrix rows record the hidden states of prefixes preceding each token).}
\textit{For every token $\mathrm{tok}$, the aggregated gradient residual
satisfies}
\[
-o_{\mathrm{tok}}
=
\underbrace{\sum_{\forall x \text{ s.t. } y_x = \mathrm{tok}} \alpha_x h_x}
_{\text{ prefixes followed by } \mathrm{tok}}
-
\underbrace{\sum_{\forall x} \alpha_x
p_\theta(\mathrm{tok}\mid x)\,h_x}
_{\text{background over all prefixes}},
\]
\textit{where $o_{\mathrm{tok}}$ is the row for $\mathrm{tok}$ in the residual for $W_{\mathrm{out}}$, $y_x$ is the token that follows prefix $x$ in a training sentence, $h_x$ is its final hidden state, and $\alpha_x>0$ are the aggregation weights. In words, $-o_{\mathrm{tok}}$ is the sum of the hidden states of the prefixes that $\mathrm{tok}$ followed, minus a background term shared by all rows.}

\begin{proof}
Write $w_v$ for the output row associated with token $v$.
For a prefix $x$ with next-token target $y_x$, the logits are
$z_{x,v}=w_v^\top h_x$, and the cross-entropy loss is
\[
\ell_x
=
-w_{y_x}^{\top}h_x
+\log\sum_v \exp(w_v^\top h_x).
\]
Differentiating with respect to an output row gives
\[
\frac{\partial\ell_x}{\partial w_{\mathrm{tok}}}
=
\left(
p_\theta(\mathrm{tok}\mid x)
-\mathbf{1}\{y_x=\mathrm{tok}\}
\right)h_x.
\]
Thus, gradient descent moves the correct output row in the direction
of $h_x$, while moving every other row in the opposite direction,
with strength proportional to its predicted probability.

Aggregating the row gradients with weights $\alpha_x$ and separating
the target and probability terms yields
\begin{multline}
-o_{\mathrm{tok}}
=
-\sum_x \alpha_x
\frac{\partial\ell_x}{\partial w_{\mathrm{tok}}} \\
=
\sum_{x:\,y_x=\mathrm{tok}}\alpha_x h_x
-
\sum_x\alpha_x p_\theta(\mathrm{tok}\mid x)h_x,
\end{multline}
as claimed.
\end{proof}

The corresponding signal backpropagated through the hidden state is
\[
-\frac{\partial\ell_x}{\partial h_x}
=
w_{y_x}-\sum_v p_\theta(v\mid x)w_v,
\]
which favors the correct output row over the probability-weighted
alternatives.

With tied embeddings, each row is used both as an output row and
as an input embedding, so its gradient is the sum of these two
contributions:
\[
o_{\mathrm{tok}}^{\mathrm{tied}}
=
o_{\mathrm{tok}}^{\mathrm{out}}
+
o_{\mathrm{tok}}^{\mathrm{in}}.
\]
The output contribution derived above remains present, but is mixed
with the gradient arising from occurrences of $\mathrm{tok}$ as an
input. Tokens used exclusively as output targets, such as an
end-of-sequence token used only as the final target and omitted from
the inputs, have no input-embedding contribution and therefore
retain exactly the expression in the lemma.

\section{Bag of Tokens, formally defined}
\label{app:bagoftokens}
For untied models, we get the following expression 
\[
\frac{\partial \mathcal{L}}{\partial (W_{\mathrm{in}})_{\mathrm{tok},j}}
= \sum_{i} \mathbf{1}[s_i=\mathrm{tok}]\,
\frac{\partial \mathcal{L}}{\partial e_{i,j}},
\]
which reveals whether a token appeared during training  with the help of the embedding matrix. Here, $\mathcal{L}$ is the training loss, $\mathrm{tok}$ is a vocabulary token (indexing a row of $W_{\mathrm{in}}$), $j$ indexes the embedding dimension, $s_i$ is the token and $e_i$ is its input embedding at position $i$, with $e_{i,j}$ its $j$-th coordinate, and $\mathbf{1}[\cdot]$ is the indicator function.

\section{Metric definitions}
\label{app:eval-metrics}

Let $X=\{x_i\}_{i=1}^{M}$ be the reference sequences, where
$M=cb$, and let $\hat X$ be the recovered sequences. We use
Hungarian matching to find a one-to-one assignment that
maximizes the sum of ROUGE-L F1 scores,
\[
\operatorname{ROUGE\text{-}L}_{F_1}(x,\hat x)
=\frac{2\operatorname{LCS}(x,\hat x)}{|x|+|\hat x|},
\]
where $\operatorname{LCS}$ is the length of the longest common
subsequence. We use the same assignment for all metrics.
Below, $\hat x_i$ denotes the prediction matched to $x_i$;
unmatched references receive zero credit.

Let $c_n(g;x)$ count the occurrences of an $n$-gram $g$ in $x$.
The overlap count and total number of $n$-grams are
\begin{align*}
O_n(x,\hat x)
&=\sum_g\min\{c_n(g;x),c_n(g;\hat x)\},\\
T_n(x)&=\max\{|x|-n+1,0\}.
\end{align*}

\paragraph{BLEU-2.}
Let $p_n=O_n(x,\hat x)/T_n(\hat x)$ be the clipped
$n$-gram precision. For $|\hat x|\geq 2$, we compute
\[
\operatorname{BLEU\text{-}2}(x,\hat x)
=
\exp\!\left(\min\left\{0,1-\frac{|x|}{|\hat x|}\right\}\right)
\sqrt{p_1p_2},
\]
and assign zero otherwise. BLEU-2 measures the precision of
recovered tokens and adjacent token pairs. Its brevity penalty
reduces the score of short, incomplete outputs.

\paragraph{ROUGE-2.}
We report bigram F1:
\[
\operatorname{ROUGE\text{-}2}(x,\hat x)
=\frac{2O_2(x,\hat x)}{T_2(x)+T_2(\hat x)},
\]
with zero assigned when the denominator is zero. ROUGE-2 F1
balances the correctness of recovered bigrams with coverage
of the reference, giving credit to partial reconstructions
that preserve local token order. For either overlap metric
$s$, the reported score is
\[
\overline{s}=\frac{1}{M}\sum_{i=1}^{M}s(x_i,\hat x_i).
\]

\paragraph{Exact sentence recovery.}
We report the percentage of reference sequences recovered
exactly:
\[
\operatorname{Exact}(\%)
=\frac{100}{M}\sum_{i=1}^{M}\mathbf{1}[x_i=\hat x_i].
\]
A sentence is recovered exactly if all its tokens match the
reference in the same order, with no missing or extra tokens.
We use this metric to measure how many private sequences
are recovered in full.

\subsection{Poor recovery from baselines}
\label{sec:poor-recovery}
A correct prefix missing only the final token receives no
exact recovery credit. DAGER reports this limitation under
next-token prediction \cite{petrov2024dager}. The final token
acts only as a target: its input position has no next-token
prediction loss, so its representation does not contribute
to the attention-gradient span used by DAGER. Such prefixes
can still obtain high overlap scores. We therefore also
report BLEU-2 and ROUGE-2 to measure partial recovery.

FILM primarily targets recovery of a single sentence from
a batch and relies heavily on memorization in the language
model \cite{gupta2022recovering}. To recover more sentences,
it repeats the search on the same gradient and penalizes
previously generated $n$-grams. This encourages different
outputs but does not ensure that each output recovers a
different sentence from the batch. Under one-to-one matching,
several candidates cannot receive credit against the same
reference. Each repetition also requires another beam search
and refinement, with gradient computations used to score
candidate edits. We discuss computational cost in
Section~\ref{sec:efficiency}.

Figures~\ref{fig:bleu2_enron}, \ref{fig:bleu2_sst2},
and~\ref{fig:bleu2_pm} show the BLEU-2 results.
Figures~\ref{fig:rouge2_enron}, \ref{fig:rouge2_sst2},
and~\ref{fig:rouge2_pm} show the ROUGE-2 results.

\section{Additional Experiments}
\subsection{Experimental configurations}
\label{app:exp-configs}
We define the $(c, b, s)$ configurations in this section and use them to analyze recovery rates for below and above the hidden dimensionality $d$.

\begin{table}[t]
    \centering
    \small
    \setlength{\tabcolsep}{4pt}
    \begin{tabular}{lrccc}
        \hline
        $(c,b,s)$ & Total tokens
        & $d=768$ & $d=1024$ & $d=4096$ \\
        \hline
        $(4,8,8)$     & 256    & Below & Below & Below \\
        $(4,16,16)$   & 1,024  & Above & At    & Below \\
        $(4,32,16)$   & 2,048  & Above & Above & Below \\
        $(8,64,16)$   & 8,192  & Above & Above & Above \\
        $(16,64,8)$   & 8,192  & Above & Above & Above \\
        $(16,64,16)$  & 16,384 & Above & Above & Above \\
        $(32,64,16)$  & 32,768 & Above & Above & Above \\
        $(32,128,16)$ & 65,536 & Above & Above & Above \\
        \hline
    \end{tabular}
    \caption{Experimental configurations and their position relative to the hidden-dimension wall. Total tokens are computed as $cbs$. The dimensions $768$, $1024$, and $4096$ correspond to \texttt{Pythia-160M}, \texttt{Pythia-410M}, and \texttt{GPT-J-6B}, respectively.}
    \label{tab:experimental_configs}
\end{table}

\subsection{Other architectural results}
\label{app:non-parallel-evals}
The token-direction decomposition used by our full attack relies on parallel-residual blocks. We assess transfer beyond the main Pythia and GPT-J evaluation using sequential-residual architectures and models with
tied embeddings. For the sequential models, we use a reduced variant that combines unembedding gradient cosine similarity, and language-model fluency, with the MLP evidence terms disabled.

\paragraph{\texttt{GPT-2.}}
\texttt{GPT-2} combines sequential residual blocks with tied embeddings ($d=768$). Across the evaluated datasets and configurations, the reduced variant reconstructs 38 sentences exactly among 57 accepted candidates,
giving a pooled sentence-level precision of $66.7\%$
(Table \ref{tab:arch-gpt2}). Acceptance is therefore less reliable than in the main experiments: for example, only 10 of 18 accepted Enron sentences are exact at $(c,b,s)=(4,16,16)$. No sentences are accepted at $(8,64,16)$ on any dataset. These results establish recovery on \texttt{GPT-2}, but the simultaneous changes to architecture and attack components prevent attributing the degradation to embedding tying alone.

\begin{table}[t]
    \centering
    \small
    \caption{GPT-2 recovery using the reduced attack variant. Each entry gives the number of precise reconstructions, followed by the number
    of accepted candidates in parentheses.}
    \label{tab:arch-gpt2}
    \setlength{\tabcolsep}{3pt}
    \begin{tabular}{@{}lcccc@{}}
        \toprule
        Dataset & $(4,8,8)$ & $(4,16,16)$ & $(4,32,16)$ & $(8,64,16)$ \\
        \midrule
        SST-2  & 5 (5)   & 2 (8)   & 1 (5)  & 0 (0) \\
        Enron  & 4 (5)   & 10 (18) & 3 (3)  & 0 (0) \\
        PubMed & 6 (6)   & 5 (5)   & 2 (2)  & 0 (0) \\
        \bottomrule
    \end{tabular}
\end{table}

\paragraph{Larger models.}
A validation run on Mistral-7B-v0.3, which uses sequential residual blocks and untied embeddings ($d=4096$), also recovers 7 SST-2 sentences at $(4,8,8)$. This provides evidence of transfer to a larger sequential model.

\subsection{Scalability and efficiency}
\label{sec:efficiency}
\begin{figure*}
    \centering
    \includegraphics[width=\linewidth]{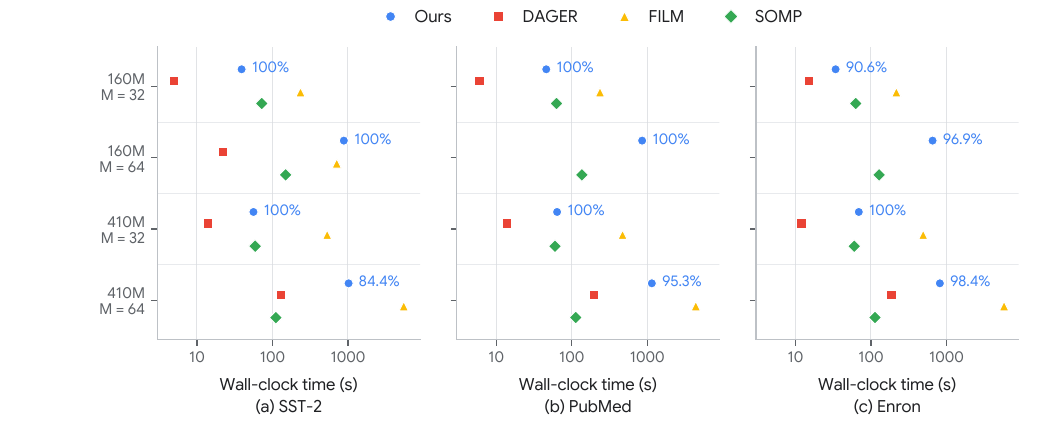}
    \caption{Wall-clock runtime and exact sentence recovery for \texttt{Pythia-160M} and \texttt{Pythia-410M}. Blue annotations give our exact recovery
    percentages, while all baselines recover 0\% exactly.}
    \label{fig:runtime}
\end{figure*}

In our pipeline, the attack cost is dominated by candidate generation and verification across peeling passes. Figure~\ref{fig:runtime} reports wall-clock times
for \texttt{Pythia-160M} and \texttt{Pythia-410M} on three datasets, using $(c,b,s)=(4,8,8)$ and $(4,16,16)$. The latter has $cbs=1024$, placing it above the wall for Pythia-160M ($d=768$) and at the wall for Pythia-410M ($d=1024$).

Our method recovers $90.6$--$100\%$ of sentences in $34$--$69$ seconds for the smaller configuration, and $84.4$--$100\%$ in $655$--$1152$ seconds for the larger one. All reported baseline runs recover zero exact sentences. These measurements establish the runtime at which our method achieves exact recovery in these settings. 

\subsection{Hyperparameter sensitivities}
\label{app:hyperparam-sensitivities}
\begin{figure}
    \centering
    \includegraphics[width=\linewidth]{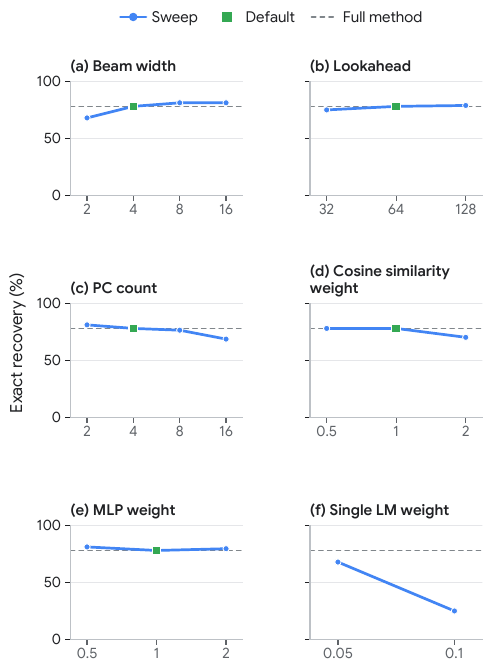}
    \caption{Ablation studies on all hyperparameters we utilize in our recovery method.}
    \label{fig:abl-hp}
\end{figure}

Figure \ref{fig:abl-hp} describes the different hyperparameters and its sensitivity with the recovery rate. Increasing beam width from two to four raises recovery from 87 to 100 sentences; beam widths of eight and sixteen both recover 104. Fluency weighting is also significant, using a single weight of $0.1$ reduces recovery to 32 sentences.

\subsection{Aggregation algorithms}
\label{app:aggregation-algos}
We compare equal averaging (\textsc{FedSGD}), size-weighted averaging, and \textsc{FedAdam} on \texttt{Pythia-160M} and SST-2 with $(c,b,s)=(4,32,16)$, one local step, $\eta=10^{-2}$.

\begin{figure}
    \centering
    \includegraphics[width=\linewidth]{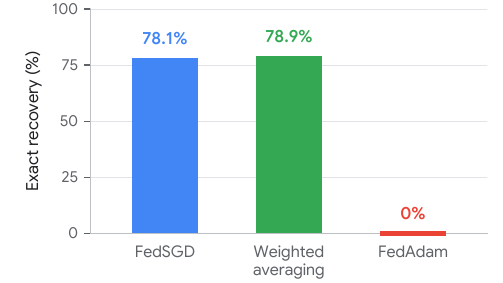}
    \caption{Comparison of aggregation algorithms on gradient recovery.}
    \label{fig:fed_agg}
\end{figure}

As seen in Figure \ref{fig:fed_agg}, \textsc{FedSGD} and weighted averaging (\textsc{FedAvg}) recover $100/128$ ($78.1\%$) and $101/128$ ($78.9\%$) sentences exactly, respectively, with no false accepts. Both preserve a linear mixture of client gradients, although weighting changes their relative contributions. On the other hand, \textsc{FedAdam} yields no exact recoveries. In our first-round setting, the second-moment accumulator starts at zero, giving the coordinatewise denominator
\begin{equation}
\sqrt{v_j}+\epsilon
=\sqrt{1-\beta_2}\,|g_j|+\epsilon,
\end{equation}
where $g$ is the aggregate gradient, $\beta_2=0.999$, and $\epsilon=10^{-8}$. When the first term dominates, the update becomes approximately proportional to $\mathrm{sign}(g_j)$, compressing relative gradient magnitudes. 
The result establishes failure of the current attack under this configuration; it does not establish protection against recovery methods that account for the server optimizer.

\subsection{Numerical precision}
\label{app:precision-numerical}
Recovery runs in FP32 by default. We compare FP32 and TF32 model loading on \texttt{Pythia-160M}, SST-2 at $(4,32,16)$, as shown in Table~\ref{tab:numerics}. Our results indicate that the attack is robust to reduced-precision matmuls.

\begin{table}\centering\small
\caption{Exact recovery across numerical-precision modes (Pythia-160M / SST-2 / $(4,32,16)$, out of $128$). FP32 and TF32 agree within seed noise.}
\label{tab:numerics}
\begin{tabular}{lc}\toprule
Mode & Recovery Rate \\ \midrule
FP32 & 100 \\
TF32 & 103 \\
\bottomrule
\end{tabular}
\end{table}

\subsection{Whitening is anomalous}
\label{app:whitening-strange}
% Soham: whitening anomalies, covariance investigations

We notice that whitening components have mixed effects in \texttt{Pythia-160M} with the configuration $(4, 32, 16)$. (Table \ref{tab:white}). In some datasets such as SST-2, which has several phrases repeated, whitening tends to harm the pipeline and causes several sentences not to be recovered. However, in other private datasets which has lesser \emph{multiplicity}, whitening has consistently improved our recovery. We leave a systematic study of this relationship to future work.

\begin{table}[h]
\centering
\caption{Whitening ablation on \texttt{Pythia-160M} (exact recovery). Full whitening dominates on the hardest above-the-wall cells.}
\label{tab:white}
\small
\setlength{\tabcolsep}{5pt}
\begin{tabular}{@{}lccc@{}}
\toprule
Whitening & SST-2 & Enron & PubMed \\
\midrule
\multicolumn{4}{@{}l}{\textit{$128\!:\!16$}} \\
full          & 100 & \textbf{123} & \textbf{117} \\
no whitening  & \textbf{111} & 52  & 29  \\
\addlinespace[2pt]
\multicolumn{4}{@{}l}{\textit{$512\!:\!16$}} \\
full          & \textbf{42} & \textbf{48} & \textbf{76} \\
no whitening & 4  & 0  & 4  \\
\bottomrule
\end{tabular}
\end{table}

\section{Implementation details}
\label{app:other-configs}
Table~\ref{tab:hyperparams} lists the hyperparameters used in all default runs.

\paragraph{Compute information.} All recovery methods were executed using a compute node equipped with one single NVIDIA A100 (40GB) GPU.

\paragraph{Statistical significance.} All experiments were repeated three times with random seeds $7$, $42$, and $67$, and we report the median result across these runs throughout this work.

\subsection{Baseline information}
\label{app:baseline-info}
We use hungarian matching with the ground truth for all baselines in order to maintain a fair comparison. 

\paragraph{DAGER.}
We do not apply reduced-rank truncation in the main experiments. Separate tests with the $d-20$ restriction on a subset of unsupported cases produced no candidates. Each search is capped at $10^7$ candidate expansions; runs reaching this limit are recorded as zero recovery, and incomplete candidates are discarded.

\paragraph{FILM.}
We reconstruct $M$ candidate sequences from a fixed gradient, accumulating bigram bans across repetitions to discourage duplicate outputs. We use pretrained checkpoints without additional training
on private data, so our evaluation does not capture the memorization benefits reported in the original paper. Starting tokens are ranked by model probability within the observed support, without an uppercase constraint.

\paragraph{SOMP.}
As no author-provided implementation was available to us, we reimplement SOMP following the algorithm and implementation details reported in the paper.

\begin{table}\centering\small
\caption{Default hyperparameters in our recovery method.}
\label{tab:hyperparams}
\begin{tabular}{@{}ll@{}}\toprule
Parameter & Value \\ \midrule
\multicolumn{2}{@{}l}{\textit{Candidate generation}} \\
Beam width                      & 4 \\
Seed batch                      & 16 \\
Lookahead                       & 64 \\
\addlinespace[2pt]
\multicolumn{2}{@{}l}{\textit{Scoring weights}} \\
Head (ordering)                 & 1.0 \\
LM fluency                      & union of $\{0.02,0.03,0.04,0.05\}$ \\
\addlinespace[2pt]
\multicolumn{2}{@{}l}{\textit{Whitening}} \\
Principal components removed    & 4 \\
Full covariance                 & yes \\
Column scaling / centering      & yes / yes \\
\addlinespace[2pt]
\multicolumn{2}{@{}l}{\textit{Verification}} \\
Min.\ anchors                   & 3 \\
End window                      & 0 \\
Input / source tolerance        & $5\times10^{-3}$ / $5\times10^{-3}$ \\
Beta / cleanup tolerance         & $2\times10^{-3}$ / $3\times10^{-3}$ \\
Source \& input anchors          & yes \\
\addlinespace[2pt]
\multicolumn{2}{@{}l}{\textit{Peeling \& budget}} \\
Peel passes                     & 24 \\
Budget / pass                   & $50{,}000$ \\
\addlinespace[2pt]
\multicolumn{2}{@{}l}{\textit{Seeds}} \\
Seeds                      & 7, 42, 67 \\
\bottomrule
\end{tabular}
\end{table}

\section{Why recovery degrades with aggregation}
\label{app:moving-the-wall}
DAGER's attention-gradient span test loses selectivity when the observed gradient spans $\mathbb{R}^{d}$~\citep{petrov2024dager}.
Our token-direction solve instead uses the first MLP's expanded feature space, whose dimensionality is $m=d_{\mathrm{MLP}}$. The resulting capacity depends on the number of distinct tokens, rather than directly on the total number of token occurrences $C$.

Let $q=|\mathcal{B}|$ denote the number of distinct tokens. As shown in the proof of Lemma \ref{lem:tokdir} in Appendix~\ref{app:tokdir}, the token directions are uniquely recovered when the first-MLP activations of the distinct tokens are linearly independent. Since these activations lie in $\mathbb{R}^{m}$, this requires $q\leq m$: the number of distinct tokens must not exceed the MLP width. Here $A=[a_{\mathrm{tok}}]_{\mathrm{tok}\in\mathcal{B}}\in\mathbb{R}^{m\times q}$ denotes the known activations and $D=[d_{\mathrm{tok}}]_{\mathrm{tok}\in\mathcal{B}}$ the unknown token directions, so that $G_{\mathrm{mlp},0}=DA^{\top}$.

The nonlinear MLP features can support $q>d$ independent columns, up to $q=m$. This does not require $G_{\mathrm{mlp},0}$ itself to have rank $q$: recovery uses the known matrix $A$. Conversely, when $q>m$, there exists a nonzero $v\in\ker A$. For any $u\in\mathbb{R}^{d}$, $(D+uv^{\top})A^{\top}=G_{\mathrm{mlp},0}$, so this linear system cannot uniquely determine all token directions.

This shifts the dimensional constraint to the MLP width, but does not establish a sharp sentence-recovery threshold at $C=m$. The condition is on the number of distinct tokens $q\leq C$, which can remain below $m$ even when $C$ exceeds it, and $q\leq m$ is necessary but not sufficient. 

Even for $q\leq m$, dependent or poorly conditioned activation columns can prevent reliable separation. Repeated tokens also merge multiple contexts into each $d_{\mathrm{tok}}$, leaving their occurrences and ordering unresolved. Thus, both support size and token multiplicity can limit recovery as aggregation grows.

Peeling reduces these mixtures by subtracting verified sentence gradients. It reduces $q$ only when all remaining occurrences of a token are removed. If no sentences are verified, however, peeling stalls and cannot simplify the residual further.

\section{Qualitative examples of recovery}
\label{app:qualitative}

We present some qualitative examples to demonstrate the kinds of sensitive sentences our method can recover in Figure \ref{fig:qualitative-leak}.

\section{Ethical Concerns \& Reproducibility}
\paragraph{Ethical Concerns.}
Our attack exposes a privacy risk in FL, a participating client may reconstruct other clients' training text from aggregate updates. This capability poses a dual-use risk, as it could enable unauthorized access to sensitive information. Our evaluation uses existing research datasets rather than data collected from deployed federated systems. Our aim is to inform privacy assessments and the development of stronger defenses, not to facilitate unauthorized data recovery.

\paragraph{Code Reproducibility.}
Our implementation and experiment configurations are
available at \url{https://github.com/DaSH-Lab-CSIS/client-stealing} to support reproducibility. The experimental setup is described in \S\ref{sec:setup}, with implementation details and default hyperparameters provided in Appendix~\ref{app:other-configs}.

\newpage

\begin{figure*}
\centering
\begin{tcolorbox}[
colback=white!95!black,
colframe=black,
fontupper=\ttfamily,
width=\textwidth,
sharp corners,
boxrule=0.5pt,
arc=0mm,
boxsep=2mm,
left=2mm,
right=2mm
]
{\normalfont\bfseries Verbatim sentences reconstructed from a single aggregated gradient}\\[1pt]

{\normalfont\bfseries Personally identifying information}\\[2pt]
john lavorato 6437 Mercer St. Houston TX 77005 Work 713\dots\\[1pt]
Kayla Harmon Enron Volume Management 713-345-1778\\[1pt]
Sure. Chris Hill (212) 946-8887 -----Original Message-----\\[6pt]

{\normalfont\bfseries Financial information}\\[2pt]
my 2001 budget proposal, as cleaned up by accounting. As it does contain salary\dots\\[1pt]
It appears that this deal, once complete will be approximately \$80k positive.\\[1pt]
Rob How good is that credit report from Bradford. Can I use it\dots\\[6pt]

{\normalfont\bfseries Medical and PII}\\[2pt]
Thanks. I went to see Dr. Shelley Leeds Rickets this am.\\[1pt]
Fine - how are you? Baby not turning over yet because he's big!!!\\[6pt]

{\normalfont\bfseries Legally privileged / confidential}\\[2pt]
This e-mail is confidential and may well also be legally privileged. If you\dots\\[1pt]
Thanks Tana - Cindy please could you put a Power of Attorney in place
\end{tcolorbox}
\caption{Qualitative examples recovered from our method. From one averaged \textsc{FedAvg} update over $128$ private Enron sentences on \texttt{Pythia-160M}, the attack reconstructs private information. Names and numbers are reproduced exactly as they appear in the training text.}
\label{fig:qualitative-leak}
\end{figure*}

\end{document}